\documentclass{article}
\usepackage[utf8]{inputenc}
\usepackage[a4paper, margin=.6in]{geometry}

\usepackage{graphicx} % Required for inserting images
\usepackage{xcolor}
\usepackage{booktabs}
\usepackage{amsmath, amsfonts}
\usepackage{adjustbox}
\usepackage{csquotes}
\usepackage[style=vancouver,sorting=nyt]{biblatex}
\usepackage{tikz}
\usepackage{hyperref}
\usepackage{longtable}
\usetikzlibrary{positioning,calc,fit,shapes,arrows.meta,backgrounds,decorations.pathreplacing,calligraphy}
\usetikzlibrary{decorations.pathmorphing} 
\usepackage{fontawesome5}
\usepackage{bm}
\usepackage{xspace}
\usepackage{numprint}

\usepackage{pgfplots}
\DeclareUnicodeCharacter{2212}{−}
\usepgfplotslibrary{groupplots,dateplot}
\usetikzlibrary{patterns,shapes.arrows}
\pgfplotsset{compat=newest}

\usepackage{verbatim}

\newcommand{\ERGS}[1]{REAS\xspace}

\title{Validating LLM-Based Essay Scorers: A Comparative Analysis in Critical Online Reasoning in Medical Education}

\author{
    \textbf{Daniel Baumartz, Alexander Mehler, Mevlüt Bagci, Maxim Konca}\\
    Goethe University Frankfurt, Germany\\
    \texttt{\{baumartz, mehler, bagci, konca\}@em.uni-frankfurt.de}
}

\date{}

\begin{document}

\maketitle

\section*{Abstract}
\label{sec:abstract}

\subsection*{Background}

Automated essay scorers (AES) based on large language models (LLMs) promise human-level accuracy. However, training, fine-tuning, and evaluating LLMs for specific assessment tasks is costly. In particular, the question arises whether these AES should be trained for specific domains or whether models based on generic LLMs might be sufficient. Human annotations, by contrast, are often more valid but far more expensive.

\subsection*{Objective}

This study evaluates whether ensembles of AES systems based on general-purpose LLMs can approximate human ratings of medical-education writing tasks by leveraging encoded linguistic and conceptual knowledge. The goal is to assess whether general-purpose models are applicable for assisting or potentially replacing human annotation in this domain.

\subsection*{Methods}

We experiment with 11 AES systems and 8 different LLMs and evaluate 668 student responses to 12 tasks, using \numprint{2076} evaluations by 7 human annotators.
These ratings are aggregated to create consistent performance levels that can be used as target classes for classifiers trained on AES outputs.

\subsection*{Results and Conclusions}

Our results show that learner classes based on human ratings can be accurately reconstructed using a linear SVM trained on a small set of AES features. A single classifier generalized well across tasks and domains, with reasoning and content features being most predictive. This suggests that out-of-the-box LLMs, even without domain-specific fine-tuning, can support automatic essay grading in medical education. While individual AES systems are insufficient alone, their combination approximates human evaluators, pointing to a reduced need for manual scoring or its supplementation with automated methods.

\section*{Lay Summary}
\label{sec:lay_summary}

\subsection*{What is currently known about this topic?}
\begin{itemize}
    \item Automated essay scoring (AES) systems based on large language models (LLMs) are increasingly seen as alternatives to human scoring.
    \item Training and fine-tuning LLM-based AES is expensive and time-consuming, suggesting the need for general-purpose LLMs in AES.
\end{itemize}

\subsection*{What does this paper add?}
\begin{itemize}
    \item Using real-world scoring data from student responses in medical education, we evaluate eleven AES systems in combination with eight LLMs to test whether general-purpose LLMs are sufficient for domain-specific AES.
    \item We show that a small and inexpensive classifier that is trained on the outputs of these LLMs and AES systems is able to replace human raters.
\end{itemize}

\subsection*{Implications for practice/or policy}
\begin{itemize}
    \item Learner classes defined by aggregated human ratings can be nearly perfectly reconstructed using a classifier trained on a small set of AES-based features.
    \item Our approach of combining multiple AES systems reduces potential bias and reliance on single models.
    \item LLMs that are not fine-tuned with domain-specific knowledge still contain enough information to enable automatic essay grading in the medical domain.
    \item Our results suggest that the need for manual scoring can be reduced and that scoring could become an automatable task. AES systems can significantly accelerate the grading process and partially replace human ratings.
\end{itemize}

\section*{Keywords}
Medical Education,
Automated Essay Scoring,
Scoring Methods,
Natural Language Processing,
Supervised Machine Learning,
Large Language Models

\section{Introduction}\label{sec:introduction}

Large Language Models (LLMs) have accelerated the automation of scoring student essays \cite{Pack:et:al:2024,Sreedhar:et:al:2025,Li:Liu:2024,Sessler:et:al:2025}.
\textit{Automated Essay Scorers} (AES) \cite{Shermis:Burstein:2003} now operate mostly end-to-end \cite{Tate:et:al:2024,Liew:Tan:2025,Wang:et:al:2025,Su:et:al:2025b}, using unprocessed text from statements and task descriptions to generate output values, such as grades and confidence scores.
Due to the large amount of data used to train LLMs, AES have substantial potential for cross-domain use \cite{Jiang:Gao:et:al:2023}. 
This flexibility raises the question of whether cross-domain AES can approximate domain-specific expert ratings and thereby reduce the need for specialised human judgments.
The present study examines this question using data from the \textit{Critical Online Reasoning in Higher Education} (CORE) project on critical online reasoning in medical education.

CORE investigates students' critical online reasoning \cite{Kosmyna:et:al:2025} when completing tasks that require the use of online sources \cite{Nagel:2025,Hartig:et:al:2026}.
It conceptualises online reasoning as comprising three interrelated facets: 
\textit{online information acquisition} (OIA), \textit{critical information evaluation} (CIE), and \textit{reasoning with evidence, argumentation, and synthesis} (REAS)~\cite{Molerov:et:al:2020,Nagel:et:al:2022}.
According to this model, students search for, assess, and integrate information into their argumentation, ultimately incorporating it into their written task responses.
As part of CORE, students of medicine and physics were instructed to investigate medical topics online and compose written statements summarizing their findings.
These statements operationalise the \ERGS{} facet.
In their tasks, the students work on four medical scenarios within the medical part of CORE:
\textit{middle ear}, \textit{respiration}, \textit{circulation}, and \textit{eye}.
Each of these topics is divided into three tasks targeting \textit{fundamental}, \textit{practical}, and \textit{transdisciplinary} reasoning, resulting in 12 scored student responses.

A gold-standard rating was established by trained doctoral students in medical education, who evaluated the students' written responses based on 160 assignment-specific criteria.
Using all these criteria, we aggregated the annotators' ratings to assign each student response to a specific performance class.
Each student, as the sole author of a response, was uniquely assigned to one class per task, producing task-specific partitions of the set of students.
We considered two variants of this partition: a \textit{binary division} distinguishing low and high performers, and a \textit{ternary division} including medium performance.
Within this framework, our \textit{objects of observation} are students’ written responses to \ERGS{} tasks from the CORE project.
The AES-generated scores constitute observed indicators derived from these responses.
We test whether performance classes, representing an operationalization of the underlying construct as derived from human annotation, can be reconstructed from these indicators.
Successful reconstruction would indicate that the AES captures distinctions similar to those made by human raters.
Treating this as a classification task allows us to adapt AES to the medical domain via Machine Learning (ML).
Our expectation is that the classifier should map student responses as accurately as possible onto the human-derived performance classes. 
If the classifier fails to do so, the scoring outputs of the underlying AES cannot be regarded as valid indicators of text quality, \textit{at least in the context of CORE}.

Rather than evaluating AES individually, we combine their outputs to capture complementary strengths, yielding higher-dimensional feature representations that give ML access to a wider range of predictive signals.
Our hypothesis is as follows:
\begin{enumerate}
    \item[\textbf{H1}] AES systems using LLMs trained on large cross-domain corpora and used without domain-specific fine-tuning encode linguistic and conceptual knowledge that enables them to approximate human assessments of student writing according to domain-specific criteria in medical education. In other words, high agreement with human ratings on medical-education tasks can be achieved without domain-specific fine-tuning.
\end{enumerate}

To test H1, we evaluate the performance of eleven AES configured with eight LLMs.
We systematically experiment with 74 combinations of these configurations, which we release as open-source components within the Docker Unified UIMA Interface (DUUI) NLP framework~\cite{Leonhardt:et:al:2023}.
The framework is distributed under the open-source AGPL-3.0 software license, while derived research data including de-identified text embeddings and instance-level class labels are made available under the CC BY-NC 4.0 license.

\subsection{Related Work}
\label{sec:related_work}
We review LLM-based AES (see \cite{Misgna:et:al:2024,Li:Ng:2024} for overviews).
We begin with medical-domain approaches.
\textcite{Gierl:et:al:2014} investigate how AES support medical assessment through scoring consistency, efficiency, and timely feedback.
They found that AES achieve human-level agreement, indicating their potential for scoring constructed responses.
\textcite{Sreedhar:et:al:2025} compare ChatGPT with faculty raters in formative medical assessments and report substantial agreement with high inter-rater reliability across rubric dimensions.
Their results suggest that LLMs can assist educators by providing consistent and scalable feedback on student performance.
\textcite{Elshama:2020} stress that medical-education assessments must align with instructional objectives and use valid, reliable tools to evaluate knowledge, skills, and attitudes.
They emphasise the need for clear reporting and the effective use of results to support ongoing curriculum improvement.
\textcite{colbertgetz:et:al:2017} propose a practical rubric for assessing usefulness in medical education, covering topics such as equivalence, the educational effect, the catalytic effect, evidence of validity, and acceptability.
These approaches show the relevance of AES for the medical field.

AES have also been studied outside the medical field.
This comprises domain-independent approaches intended to generalise across fields.
The ESSAJUDGE~\cite{Su:et:al:2025} benchmark systematically evaluates the impact of LLMs on essay scoring by assessing 18 models.
The authors propose a method to capture linguistic patterns and contextual cues across lexical, sentence, and discourse levels.
Another study on automated essay scoring is T-MES \cite{wang:liu:2025}.
It addresses two key limitations of multi-trait scoring: the lack of trait-specific representations and the neglect of relationships between traits.
T-MES uses a mixture-of-experts framework to generate trait-specific representations and introduces regularization schemes that enhance cross-trait collaboration and scoring diversity.
TOREE~\cite{zhuang:et:al:2024} is a benchmark for evaluating topic relevance in student essays.
Their two-step framework, which combines supervised fine-tuning and preference learning, improves both relevance scoring and feedback generation.
It emphasises the importance of topic adherence, which is often overlooked in AES.
\textcite{wang:et:al:2018} apply reinforcement learning to AES, modeling scoring as a rating-scheme-guided decision sequence that achieves more consistent agreement with human raters.
Their approach, which models essay scoring as a sequential decision process and incorporates rating schema information, achieves more consistent agreement with human raters than traditional supervised methods.
\textcite{Goura:et:al:2022} present an enhanced AES that integrates classical and neural features, refining the feature-extraction and modeling pipeline to better capture key dimensions of student writing.
Their work contributes to the wider AES landscape by refining the feature extraction and modelling pipeline to accommodate the different aspects of student writing.
\textcite{Bashendy:et:al:2024} advance Arabic AES by introducing a trait-annotated dataset that evaluates essay dimensions, including relevance, organization, vocabulary, style, development, and grammar.
\textcite{Yoo:et:al:2025} present DREsS, a large-scale rubric-based AES dataset for EFL writing that captures content, organization, and language-use traits, enabling models that align more closely with human grading standards.
\textcite{Gaudeau:2025} moves away from using a single gold standard label.
She proposes to learn from individual raters to account for rater variability and provide confidence estimates.
\textcite{Chakravarty:et:al:2025} model AES reliability as a classification task, treating score-release decisions as predictions.
\textcite{Li:Ng:2024} critically review the field of AES, arguing that recent research focuses on making small improvements to standard datasets. They urge the community to broaden its horizons towards understanding, generalisation across different prompts, richer trait annotations, and educational impact in the real world.

Most existing work develops new AES and evaluates them on gold-standard data, with some efforts focused on benchmarking LLMs. To address the gaps in these approaches, we adopt a more robust, integrative method.
Using real-world assessment data with human ratings, we present a classification approach that integrates multiple AES and LLMs for greater generalizability. Specifically, we combine 11 AES described in \autoref{sec:methods:essay_scorer}.

\section{Methods}
\label{sec:methods}
\begin{figure}[!htbp]
\begin{center}
\begin{adjustbox}{max width=1\linewidth, trim={3cm 0 0 0}, clip}
% \documentclass[tikz,margin=0.5pt]{standalone}
% if set to true, multiple comparison and take boxes are drawn.
% if set to false (default), these boxes are only drawn once to show that there's only one decide per cycle.
%\multidecidetrue
\definecolorset{cmyk}{GU-}{}{%
	Goethe-Blau,1,0.2,0,0.4;%
	Hellgrau,0.04,0.04,0.05,0.02;%
	Sandgrau,0.12,0.09,0.13,0;%
	Dunkelgrau,0.25,0.25,0.3,0.75;%
	Purple,0.08,1,0.3,0.36;%
	Emo-Rot,0.04,1,0.8,0.07;%
	Senfgelb,0.01,0.25,1,0.05;%
	Gruen,0.62,0.4,0.87,0.09;%
	Magenta,0.08,0.86,0.12,0.12;%
	Orange,0,0.7,1,0.04;%
	Sonnengelb,0,0.12,0.95,0;%
	Helles-Gruen,0.4,0.17,0.81,0.07;%
	Lichtblau,0.8,0,0.06,0.04}

\definecolor{HUDarkBlue}{RGB}{16,82,132}
\definecolor{HULightBlue}{RGB}{55,140,195} %% HTML: 378cc3
\definecolor{HUBlueText}{RGB}{0,115,186}
\definecolor{HUDarkGrey}{RGB}{206,207,209}
\definecolor{HUMedGrey}{RGB}{219,220,221}
\definecolor{HULightGrey}{RGB}{241,241,241} %% HTML: f1f1f1
\definecolor{HUDarkOrange}{RGB}{146,76,0} %% HTML: 924c00

\definecolor{SeminarBlau}{RGB}{0,154,224}
\definecolor{SeminarRot}{rgb}{0.75,0,0}
\definecolor{SeminarGruen}{RGB}{0,128,0}
\definecolor{SeminarOrange}{RGB}{237,167,45} % rgb: 
\renewcommand\familydefault{\sfdefault}
\newcommand{\Docker}{\textcolor{GU-Lichtblau}{\faDocker}}
\newcommand{\Database}{\textcolor{GU-Gruen}{\faDatabase}}
\newcommand{\File}{\textcolor{GU-Magenta}{\faFile}}
\newcommand{\Random}{\textcolor{GU-Purple}{\faRandom}}
\newcommand{\DatabaseRandom}{\textcolor{GU-Emo-Rot}{\faDatabase}}
\newcommand{\CloneRandom}{\textcolor{GU-Emo-Rot}{\faClone}}
\newcommand{\Clone}{\textcolor{GU-Gruen}{\faClone}}
\newcommand{\Star}{\textcolor{GU-Emo-Rot}{\faStar}}
\newcommand{\Square}{\textcolor{GU-Emo-Rot}{\faSquare}}
% \begin{document}
        % \tikzstyle{startstop} = [rectangle, rounded corners, minimum width=3cm, minimum height=1cm,text centered, draw=black, fill=red!30, font=\Huge]
        \tikzstyle{startstop} = [rectangle, rounded corners, minimum width=3cm, minimum height=1cm,text centered, draw=black, fill=red!30, font=\Huge, text width=7cm, inner sep=10pt]
        \tikzstyle{process} = [rectangle, inner sep=10pt,  minimum height=1cm, text centered, text width=7cm, draw=black, fill=orange!30, font=\Huge]
        \tikzstyle{arrow} = [thick,->,>=stealth, line width=5pt]
                \tikzset{
              Box/.style={rectangle, inner sep=5.5pt, fill=#1!30!white, draw=#1, line width=1pt, text width=2cm, align=center},
              Label/.style={circle, inner sep=1pt, line width=1pt, fill=#1!30!white, draw=#1, scale=1.2, xshift=0.3cm, yshift=0.4cm},
              Label/.default={GU-Emo-Rot!50!white},
              Box/.default={GU-Lichtblau},
              Input/.style={Box=GU-Helles-Gruen},
              Features/.style={Box=GU-Orange},
              Output/.style={Box=GU-Orange},
              InFile/.style={label={[Label=GU-Helles-Gruen, font=\footnotesize, opacity=#1, xshift=0cm]left:\textcolor{GU-Helles-Gruen}{\faFile*[regular]}}},
              InVideo/.style={label={[Label=GU-Helles-Gruen, font=\footnotesize, opacity=1, xshift=0cm]left:\textcolor{GU-Helles-Gruen}{\faFileVideo[regular]}}},
              InAudio/.style={label={[Label=GU-Helles-Gruen, font=\footnotesize, opacity=1, xshift=0cm]left:\textcolor{GU-Helles-Gruen}{\faFileAudio[regular]}}},
              InImage/.style={label={[Label=GU-Helles-Gruen, font=\footnotesize, xshift=0cm]left:\textcolor{GU-Helles-Gruen}{\faFileImage[regular]}}},
              InXMI/.style={label={[Label=GU-Goethe-Blau, font=\footnotesize, xshift=0cm]left:\faFileCode[regular]}},
              Out/.style={label={[Label=GU-Orange, font=\footnotesize, opacity=#1, xshift=0cm]left:\textcolor{GU-Orange}{\faFileCode[regular]}}},
              ML/.style={Box=GU-Lichtblau, inner sep=5pt, yshift=0.2cm, align=center, minimum height=2pt, text width=0.56cm, label={[opacity=#1, xshift=0.1cm]left:\textcolor{GU-Lichtblau}{\faCogs}}},
               ML1/.style={label={[Label=GU-Lichtblau, font=\footnotesize, opacity=#1, xshift=0cm]left:\textcolor{GU-Lichtblau}{\faCogs}}},
               TF/.style={label={[Label=GU-Helles-Gruen, font=\large, opacity=#1, xshift=0cm]left:\textcolor{GU-Helles-Gruen}{\faFile*[regular]}}},
            MLXY/.style={label={[Label=GU-Lichtblau]left:\textcolor{GU-Lichtblau}{\tiny\faCogs}}},
              Line/.style={->, line width=1.0pt, draw=SeminarOrange!80!black}
             }

    \begin{tikzpicture}
      % \begin{scope}[shift={(10cm,0)}]
       \node (LL1) [Box] {\textit{LLM1}};
       \node (LL2) [Box, below=0.5cm of LL1] {\textit{LLM2}};
       \node (LLMore) [Box, below=0.5cm of LL2] {\textbf{...}};
       \node (LL8) [Box, below=0.5cm of LLMore] {\textit{LLM8}};
       \node[fit=(LL1) (LL8), draw=GU-Goethe-Blau, line width=2pt, inner sep=20pt, label={[yshift=-1.5ex]below:{\Large LLMs}}] (LL) {};

       \node (ES1) [Box, right=4cm of LL1] {\textit{ES1}};
       \node (ES2) [Box, below=0.5cm of ES1] {\textit{ES2}};
       \node (ESMore) [Box, below=0.5cm of ES2] {\textbf{...}};
       \node (ES11) [Box, below=0.5cm of ESMore] {\textit{ES11}};
       \node[fit=(ES1) (ES11), draw=GU-Goethe-Blau, line width=2pt, inner sep=20pt, label={[yshift=-1.5ex]below:{\Large Essay Scorer}}] (ES) {};

       \node (F1) [Features, right=4cm of ES1] {\textit{F1}};
       \node (F2) [Features, below=0.5cm of F1] {\textit{F2}};
       \node (FMore) [Features, below=0.5cm of F2] {\textbf{...}};
       \node (F203) [Features, below=0.5cm of FMore] {\textit{F203}};
       \node[fit=(F1) (F203), draw=GU-Orange, line width=2pt, inner sep=20pt, label={[yshift=-1.5ex]below:{\Large Features}}] (F) {};

       % \node (Task) [Box, above=1.45cm of F1] {Task \& Statements};
       \node (Task) [Box, above=1.45cm of F1, xshift=-4.6cm] {Tasks \& Statements};
       
       \node (Text1) [TF=1, above=0.6cm of ES, xshift=-0.2cm] {};
       \node (Text2) [TF=0.8, above=0.85cm of ES, xshift=0.15cm] {}; 
       \node (Text3) [TF=0.7, above=1.1cm of ES, xshift=0.5cm] {};
       \node (Text4) [TF=0.6, above=1.35cm of ES, xshift=0.85cm] {};

       \node (evocls) [Box, right=4cm of F2] {Evolutionary Search};
       \node (classifier) [Box, below=0.5cm of evocls] {Classifier};
       \node[fit=(evocls) (classifier), draw=GU-Goethe-Blau, line width=2pt, inner sep=20pt, label={[yshift=-1.5ex]below:{}}] (evoclass) {};

       \node (shap) [Box, below=1.5cm of classifier] {SHAP Analysis};
       \node (distribution) [Box, above=1.5cm of evocls] {Distribution Analysis};

       % \node (binning) [Box, right=8cm of Task] {Binning 2/3 classes};
       \node (binning) [Box, right=12.6cm of Task] {Binning 2/3 classes};

       \node (performance) [Box, below=3.775cm of binning] {Performance Prediction};

        \begin{scope}[on background layer]
           	% \draw [Line,dashed] (LL1)--(ES1.west);
            % \draw [Line,dashed] (LL1)--(ES2.west);
            \draw [Line,dashed] (LL1.east)--(ESMore.west);
            \draw [Line,dashed] (LL1.east)--(ES11.west);
    
            % \draw [Line,dashed] (LL2)--(ES1.west);
            % \draw [Line,dashed] (LL2)--(ES2.west);
            \draw [Line,dashed] (LL2.east)--(ESMore.west);
            \draw [Line,dashed] (LL2.east)--(ES11.west);
    
            % \draw [Line,dashed] (LLMore)--(ES1.west);
            % \draw [Line,dashed] (LLMore)--(ES2.west);
            \draw [Line,dashed] (LLMore.east)--(ESMore.west);
            \draw [Line,dashed] (LLMore.east)--(ES11.west);
    
            % \draw [Line,dashed] (LL8)--(ES1.west);
            % \draw [Line,dashed] (LL8)--(ES2.west);
            \draw [Line,dashed] (LL8.east)--(ESMore.west);
            \draw [Line,dashed] (LL8.east)--(ES11.west);
    
            \draw [Line,dashed] ($(Text1.north)+(-0.15cm,0)$)--($(ES1.north)+(-0.35cm,0)$);
            \draw [Line,dashed] ($(Text2.north)+(-0.15cm,0)$)--($(ES2.north)+(-0.0cm,0)$);
            \draw [Line,dashed] ($(Text3.north)+(-0.15cm,0)$)--($(ESMore.north)+(0.35cm,0)$);
            % \draw [Line,dashed] (Text2)--(ES2);
            % \draw [Line,dashed] (Text3)--(ESMore);
            \draw [Line,dashed] ($(Text4.north)+(-0.15cm,0)$)--($(ES11.north)+(0.7cm,0)$);
            % \draw [Line,dashed] (Text4)--(ES11);
            \draw [Line] (ES.east)--(F.west);
            
            \draw [Line] (F.east)-|+(1.5,2)|-(distribution.west);
            \draw [Line] (F.east)--($(evoclass.west)+(0cm,0.1cm)$);
            \draw [Line] (F.east)-|+(1.5,-2)|-(shap.west);
            \draw [Line] (evoclass.south)--(shap.north);
    
            \draw [Line] ($(evocls.south)+(-0.25cm,0cm)$)--($(classifier.north)+(-0.25cm,0cm)$);
            \draw [Line] ($(classifier.north)+(0.25cm,0cm)$)--($(evocls.south)+(0.25cm,0cm)$);
    
            \draw [Line,dashed] ($(classifier.east)+(0.0cm,0.1cm)$)--($(performance.west)+(0.0cm,0.05cm)$);
            \draw [Line,dashed] ($(performance.west)+(0.0cm,-0.2cm)$)--($(classifier.east)+(0.0cm,-0.15cm)$);
    
            \draw [Line] (binning.south)--(performance.north);
    
            % \draw [Line] (Task.east)--node[midway, above]{Manual Rubrics}(binning.west);
            \draw [Line] (Task.east)--node[midway, above]{Manual Rating}(binning.west);
        \end{scope}
    % \end{scope}
    \end{tikzpicture}
% \end{document}
\end{adjustbox}
\caption{%
Analysis pipeline from LLM-based automatic essay scoring to student-performance prediction using a classifier trained on human-graded essays and feature selection via an evolutionary algorithm.
}
\label{fig:Workflow}
\end{center}
\end{figure}
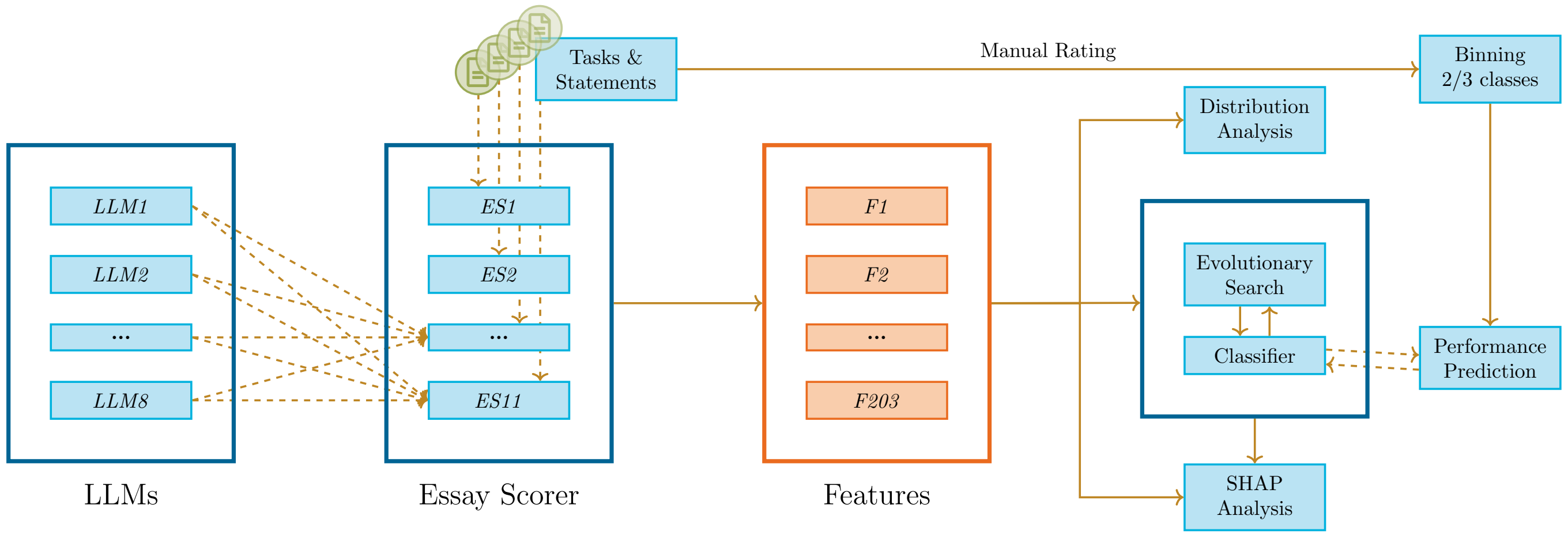

To test H1, we evaluate 11 AES on 668 manually rated student responses from the four CORE medical education assessments.
We compare the AES in \autoref{tab:essay_scorer_overview} and~\autoref{sec:methods:essay_scorer}, which output one to six scores each, yielding 35 features in total. Two systems are transformer-based~\cite{sun:wang:2024}, and nine are LLM-based.
To systematically evaluate the latter, we conduct experiments with the eight LLMs of \autoref{tab:llms_overview}. This allows us to compare how model families and parameter sizes affect grading.
See \autoref{fig:Workflow}) for an overview of the experiment workflow which is described in the remainder of this section.

\begin{table}[!htbp]
    \centering
    \begin{tabular}{l|l|c|l|l}
        \toprule
        Model & Essay Scorer Tool & \# Output Features & LLM-based prompting & Reference\\
        \midrule
        \textit{ES1} & engessay-grading-ml & 6 & transformer-based & \textcite{sun:wang:2024} \\
        \textit{ES2}  & ielts-essay-scoring  & 5  & transformer-based & \textcite{sun:wang:2024}  \\
        \textit{ES3}  & llmaesslow-scorer  & 4  & yes & \textcite{Xiao:et:al:2025} \\
        \textit{ES4}  & aes-asap-mts-scorer  & 4  & yes & \textcite{lee:et:al:2024} \\
        \textit{ES5}  & aag-scorer  & 1  & yes & \textcite{Yeung:et:al:2025} \\
        \textit{ES6}  & aes-asap-vanilla-scorer  & 1  & yes & \textcite{lee:et:al:2024} \\
        \textit{ES7}  & be-grading-scorer  & 1  & yes & \textcite{Yousef:et:al:2025} \\
        \textit{ES8}  & grading-like-human-scorer  & 4  & yes & \textcite{Xie:et:al:2024} \\
        \textit{ES9}  & llmaes-scorer-zeroshot & 4  & yes & \textcite{Xiao:et:al:2025} \\
        \textit{ES10} &  grading-medical-education  & 1  & yes & \textcite{Grevisse:2024}  \\
        \textit{ES11}  &  aes-toefel11-mts-scorer  & 4  & yes & \textcite{lee:et:al:2024} \\
        \bottomrule
    \end{tabular}
    \caption{%
    AES used in this study, along with their output features and whether or not they utilize an LLM.}
    \label{tab:essay_scorer_overview}
\end{table}

\begin{table}[!htbp]
\centering
\small
    \begin{tabular}{r|l|l|l|l|l}
        \toprule
         & Tool & Feature & Category & Prompt & Prompt Type \\
        \midrule
        F1 & ES1 & cohesion & Language & - & -\\
        F2 & ES1 & vocabulary & Language & - & -\\
        F3 & ES1 & syntax & Language & -   & -\\
        F4 & ES1 & phraseology & Language & - & - \\
        F5 & ES1 & grammar & Language & - & - \\
        F6 & ES1 & conventions & Language & - & - \\
        \midrule
        F7 & ES2 & Task Achievement & Content & - & - \\
        F8 & ES2 & Vocabulary & Language & - & - \\
        F9 & ES2 & Coherence and Cohesion & Structure & - & -\\
        F10 & ES2 & Grammar & Language & - & - \\
        F11 & ES2 & Overall & Overall & - & -\\
        \midrule
        F12 & ES3 & LLMAESSlow-ContentScore & Content & P1 & Prompt: \enquote{true\_rubrics}\\
        F13 & ES3 & LLMAESSlow-LanguageScore & Language & P1 & Prompt: \enquote{true\_rubrics}\\
        F14 & ES3 & LLMAESSlow-StructureScore & Structure & P1 & Prompt: \enquote{true\_rubrics}\\
        F15 & ES3 & LLMAESSlow-OverallScore & Overall & P1 & Prompt: \enquote{true\_rubrics}\\
        \midrule
        F16 & ES4 & Position and Thesis Clarity & Content & P2F1 & Type: \enquote{MTS}; Trait:\enquote{1}; Rubric:\enquote{1}\\
        F17 & ES4 & Supporting Details and Evidence & Content & P2F2 & Type: \enquote{MTS}; Trait:\enquote{1}; Rubric:\enquote{2}\\
        F18 & ES4 & Organization and Structure & Structure & P2F3 & Type: \enquote{MTS}; Trait:\enquote{1}; Rubric:\enquote{3}\\
        F19 & ES4 & Style, Language, and Audience & Language & P2F4 & Type: \enquote{MTS}; Trait:\enquote{1}; Rubric:\enquote{4}\\
         &  & Awareness &  &  & \\
        \midrule
        F20 & ES5 & AAGScore & Overall & P3 & Category:\enquote{AGG}\\
        \midrule
        F21 & ES6 & Vanilla & Overall & P4 & Type:\enquote{Vanilla}; Prompt ID: \enquote{1}\\
        \midrule
        F22 & ES7 & BeGradingScore & Overall & P5 & Category:\enquote{Scoring} \\
        \midrule
        F23 & ES8 & GLH-ScoringScore & Overall & P6 & Category:\enquote{Scoring}\&\enquote{Reflection} \\
        F24 & ES8 & GLH-ScoringConfidence & Confidence & P6 & Category:\enquote{Scoring}\&\enquote{Reflection} \\
        F25 & ES8 & GLH-ReflectionScore & Reflection & P6 & Category:\enquote{Scoring}\&\enquote{Reflection}\\
        F26 & ES8 & GLH-ReflectionConfidence & Confidence & P6 & Category:\enquote{Scoring}\&\enquote{Reflection}\\
        \midrule
        F27 & ES9 & LLMAESS-ContentScore & Content & P7 & Category:\enquote{zeroshot}; Rubric: \enquote{1}\\
        F28 & ES9 & LLMAESS-LanguageScore & Language & P7 & Category:\enquote{zeroshot}; Rubric: \enquote{1}\\
        F29 & ES9 & LLMAESS-StructureScore & Structure & P7 & Category:\enquote{zeroshot}; Rubric: \enquote{1} \\
        F30 & ES9 & LLMAESS-OverallScore & Overall & P7 & Category:\enquote{zeroshot}; Rubric: \enquote{1} \\
        \midrule
        F31 & ES10 & MedScore & Overall & P8 & Grading medical education\\
        \midrule
        F32 & ES11 & Task Response & Content & P9F1 & Type: \enquote{TOEFEL11}; Trait:\enquote{1}; Rubric:\enquote{1}\\
        F33 & ES11 & Coherence and Cohesion & Structure & P9F2 & Type: \enquote{TOEFEL11}; Trait:\enquote{1}; Rubric:\enquote{2}\\
        F34 & ES11 & Lexical Resource & Language & P9F3 & Type: \enquote{TOEFEL11}; Trait:\enquote{1}; Rubric:\enquote{3}\\
        F35 & ES11 & Grammatical Range and Accuracy & Language & P9F4 & Type: \enquote{TOEFEL11}; Trait:\enquote{1}; Rubric:\enquote{4} \\
        \bottomrule
    \end{tabular}
\caption{Overview of the essay scorer outputs and utilized prompts that we used as features in our classification.}
\label{tab:features_overview}
\end{table}

\begin{table}[!htbp]
    \centering
    \begin{tabular}{l|l|r|l}
        \toprule
         & LLM & Parameters & Reference \\
        \midrule
        LL3 & Llama 3.2 & 3b & \textcite{Grattafiori:et:al:2024} \\
        LQ8 & Qwen 3 & 8b & \textcite{Qwen:2025} \\
        LQ14 & Qwen 3 & 14b & \textcite{Qwen:2025} \\
        LQ32 & Qwen 3 &  32b & \textcite{Qwen:2025} \\
        LG20 & GPT-OSS & 20b & \textcite{OpenAI:2025} \\
        LG120 & GPT-OSS & 120b & \textcite{OpenAI:2025} \\
        LD70 & DeepSeek R1 & 70b & \textcite{DeepSeek:2025} \\
        LD671 & DeepSeek 3.1 & 671b & \textcite{DeepSeek:2024} \\
        \bottomrule
    \end{tabular}
    \caption{Overview of the LLMs evaluated in this study.}
    \label{tab:llms_overview}
\end{table}

\begin{table}[!htbp]
    \centering
\begin{tabular}{l|r|r|r|r|r|r|r|r}
\toprule
\textbf{Task} & \textbf{\# Criteria} & \textbf{\# Answers} & \textbf{\# Rater} & \textbf{$K\alpha_n$} & \textbf{$K\alpha_i$} & \textbf{$C\alpha_l$} & \textbf{$C\alpha_p$} & \textbf{ICC(3,k)} \\
\midrule
Respiration F & 20 & 59 & 3 & 0.54 & 0.57 & 0.84 & 0.84 & 0.84 \\
Respiration P & 19 & 45 & 3 & 0.64 & 0.69 & 0.88 & 0.88 & 0.88 \\
Respiration T & 12 & 45 & 3 & 0.45 & 0.73 & 0.91 & 0.91 & 0.91 \\
Eye F & 19 & 52 & 3 & 0.69 & 0.75 & 0.90 & 0.90 & 0.90 \\
Eye P & 15 & 50 & 4 & 0.59 & 0.64 & 0.88 & 0.88 & 0.88 \\
Eye T & 14 & 49 & 4 & 0.68 & 0.74 & 0.92 & 0.92 & 0.92 \\
Circulation F & 19 & 58 & 3 & 0.59 & 0.61 & 0.81 & 0.82 & 0.81 \\
Circulation P & 23 & 49 & 3 & 0.65 & 0.67 & 0.86 & 0.87 & 0.86 \\
Circulation T & 11 & 50 & 2 & 0.62 & 0.84 & 0.92 & 0.92 & 0.92 \\
Middle Ear F & 20 & 76 & 5 & 0.64 & 0.64 & 0.87 & 0.88 & 0.87 \\
Middle Ear P & 23 & 88 & 5 & 0.69 & 0.72 & 0.92 & 0.93 & 0.92 \\
Middle Ear T & 13 & 47 & 3 & 0.57 & 0.82 & 0.93 & 0.93 & 0.93 \\
\bottomrule
\end{tabular}
    \caption{Overview of the number of criteria, student responses and raters per CORE assessment task (F: Fundamental; P: Practical; T = Transdisciplinary), and the inter-annotator agreement calculated using mean Krippendorff's and Cronbach's alpha and intraclass correlation
    (\textit{$K\alpha_n$}: Krippendorff's alpha nominal;
    \textit{$K\alpha_i$}: Krippendorff's alpha interval;
    \textit{$C\alpha_l$}: Cronbach's alpha listwise;
    \textit{$C\alpha_p$}: Cronbach's alpha pairwise).
    }
    \label{tab:ratings}
\end{table}

We use \numprint{2076} ratings of 668 student statements from seven annotators and report inter-annotator reliability for each of the 12 CORE task-context combinations, both individually and combined, in \autoref{tab:ratings}.
As the ratings have not yet been validated, we evaluate them extensively using multiple metrics and test different binning strategies to mitigate inconsistencies.
The rating criteria fall into two categories:
\begin{enumerate}
	\item Four \textit{general criteria} apply across all CORE assessments, including medical tasks: coherence of the statement; number of sources cited with and without clear reference; and a control item indicating whether the task was answered as instructed.
	\item 160 \textit{task-specific criteria}, which vary by domain, task, and context, for example whether the student named at least one necessary examination.
\end{enumerate}
When we calculate the agreement using Krippendorff's alpha across all tasks combined, we observe moderate and average values of 64\% (nominal difference function) and 69\% (interval)~\cite{Carletta:1996,Castro:2017}.
The interclass correlation coefficients range from 81\% to 93\%, indicating excellent inter-rater reliability~\cite{Cicchetti:Sparrow:1981,Vallat:2018}.
Similarly, Cronbach's alpha estimates good to excellent 82\% to 93\% consistency when calculated using pairwise, and 81\% to 93\% using listwise deletion~\cite{George:Mallery:2016,Vallat:2018}.

Instead of using individual raters or single criteria, we aggregate all rater- and criterion-level information into a single \emph{performance class}.
Thus, we assign one performance score to every student response.
This is done in five steps:
\begin{enumerate}
    \item We exclude responses for which raters did not confirm that the student engaged seriously with the assignment.
\item The raw ratings are normalised to 0 to 1, with 1 representing the more positive criterion value (e.g. corresponding to the \enquote{yes} answer in a Boolean \enquote{yes or no} criterion).
\item We compute the mean of the normalised criteria scores for each rater, reducing all criteria to a single score in which higher values indicate better performance.
\item Because the preliminary ratings have not been consolidated into a gold score, we average the criteria scores from all raters for each response, yielding a single score per student response.
\item Using \textit{KBinsDiscretizer} of \textit{scikit-learn}~\cite{Pedregosa:et:al:2011}, we then bin the scores into two or three intervals using uniform and quantile strategies.
\end{enumerate}
We generate four performance measures for each student response:
binary separation into high and low performance classes, and ternary separation into high, medium, and low performance classes.
Both splits use either a uniform or quantile strategy: the uniform split divides the score range evenly, while the quantile split assigns an equal number of students to each bin.
This allows the levels of the performance classes to be chosen independently of the underlying rating criteria and scheme.

Next, we consider the automatic assignment of students to performance classes, that is, the reproduction of the latter two separations, as a classification task. Specifically, the task is to predict the performance classes of students based on features of their responses that are not annotated  by human experts, but by the AES evaluated here.
We treat the AES outputs as features and train a linear SVM (scikit-learn’s \textit{LinearSVC}) to predict the performance classes.
All 668 responses are treated as instances of a single classification task; we train a classifier across all assessments rather than task-specific models.
We use evolutionary feature selection \cite{Fortin:et:al:2012} to identify the AES features that contribute most to classifier performance.
We use a stratified, shuffled split of 10 folds, with a train/test size of 80/20\% and let the search work with a population of 5000 for 100 generations.
In this way, the search deploys an evolutionary algorithm that will select and remove features, trains a classifier based on these feature selection subset, measures the performance by means of Macro F1 scores and finally produces an optimal set of features that yields the highest performance.
We limit the number of features that the search may select to analyze how classifier performance varies with feature count and use the following groups of features:

\begin{enumerate}
    \item \textbf{FG1} \textit{All features} comprising all AES/LLM combinations for a total of 203 features.
    \item \textbf{FG2} \textit{Per-LLM features}, that is, the outputs of each individual LLM together with two non-LLM tools. We examine eight combinations, with 35 features each.
    \item \textbf{FG3} \textit{Per-family features} by grouping LLMs by family: Two DeepSeek and two GPT-OSS models (59 features each), and three Qwen-based models (83 features).
\end{enumerate}

Using these features, we train SVM classifiers to predict student performance classes from the AES scores of the written responses.

\subsection{Essay Scorer}\label{sec:methods:essay_scorer}
We describe the two transformer- and nine LLM-based AES evaluated in this study (see \autoref{tab:features_overview} for an overview and their outputs):
\begin{enumerate}
\item \textbf{\textit{ES1}, \textit{ES2}} \cite{sun:wang:2024} are transformer-based systems.
\textit{engessay-grading-ml}~\cite{sun:wang:2024} is a transformer-based system that produces six language-proficiency scores to assess  cohesion, syntactic structure, and grammatical accuracy.
It is a fine-tuned RoBERTa model trained on the ELLIPSE corpus, which was manually scored by expert raters.
\textit{ielts-essay-scoring} is also transformer-based; trained on the IELTS corpus of exam essays \cite{sun:wang:2024}, it grades aspects such as task achievement and grammar and provides a composite score.
\item \textbf{\textit{ES3}, \textit{ES9}} \cite{Xiao:et:al:2025} grade essays in two ways: (1) the LLM produces numerical scores and (2) generates score justifications. This yields four outputs: three category scores (content, language, structure) and an overall score formed by their sum.
We apply this approach via \textit{llmaesslow-scorer} and \textit{llmaes-scorer-zeroshot}.

\item \textbf{\textit{ES5}} Prompt engineering enables AES optimization without fine-tuning, as shown by \textcite{Yeung:et:al:2025}, who use zero-shot learning to develop \textit{Automated Assignment Grading} (\textit{aag-scorer}); it provides scores from 1--10 with explanatory feedback.

\item \textbf{\textit{ES10}} 
\textit{grading-medical-education} \cite{Grevisse:2024} uses zero-shot prompting on undergraduate medical-education data.
The prompt uses the LLM to provide a grade based on the given question and the student's response. There are no specific instructions or information about the medicine domain, except that the LLM is instructed to act as \enquote{medical teacher} in the system prompt. In our study, we replaced this term with the more generalised term \enquote{teacher}, meaning the prompt could also be used in other fields. The other prompts in our list of tools are also indifferent to the applied domain.

\item \textbf{\textit{ES8}} \textit{grading-like-human-scorer} \cite{Xie:et:al:2024} uses a zero-shot approach in which the LLM grades an essay, then reconsiders its decision, yielding a primary and a reflected score along with confidence values.

\item \textbf{\textit{ES4}, \textit{ES6}, \textit{ES11}} The \textit{Multi-Trait Specialisation} (MTS) framework \cite{lee:et:al:2024} uses zero-shot prompting to generate rubric-based criteria and score writing traits, evaluated on TOEFL11 and ASAP and compared with a vanilla overall-score setup. We use MTS via \textit{aes-asap-mts-scorer} (\textit{ES4}) and \textit{aes-toefel11-mts-scorer} (\textit{ES11}), and the vanilla prompt via \textit{aes-asap-vanilla-scorer} (\textit{ES6}).
MTS yields four trait scores (clarity, task response, organization, grammar); the vanilla method outputs a single score.

\item \textbf{\textit{ES7}} \textit{be-grading-score} \cite{Yousef:et:al:2025} is a fine-tuned LLM to grade programming assignments. As with \textit{ES10}, we remove programming elements and replace \textit{code} with \textit{essay}. We include ES7 via its prompt setup, but do not fine-tune on any data.
\end{enumerate}

All these systems are either zero-shot across multiple LLMs or transformer-based, enabling domain-independent use and ensuring comparable AES results in our study.

\subsection{Ethical Considerations}
\label{sec:ethics}
The study uses written statements produced by participants in an educational setting, as well as rating data generated by human raters.
As part of a multicenter research project, this study was first approved by the ethics committee of LMU (approval reference no.~23-0688).  
We received the dataset in anonymised form as part of an existing educational assessment in the context of the CORE project.
Although we had access to the original written texts, all personally identifiable information had been removed before we received the data.

\section{Results}
\label{sec:results}
In \autoref{sec:results:evo_search}, we present the results for H1, evaluating how well AES outputs predict student performance and how closely these predictions align with expert assessments in medical education. Since feature selection was optimised via evolutionary search, \autoref{sec:results:distribution_analysis} analyses feature contributions and the role of each AES.

\subsection{Classification Results}
\label{sec:results:evo_search}
\begin{figure}[ht!]
    \centering
    \includegraphics[width=1\linewidth]{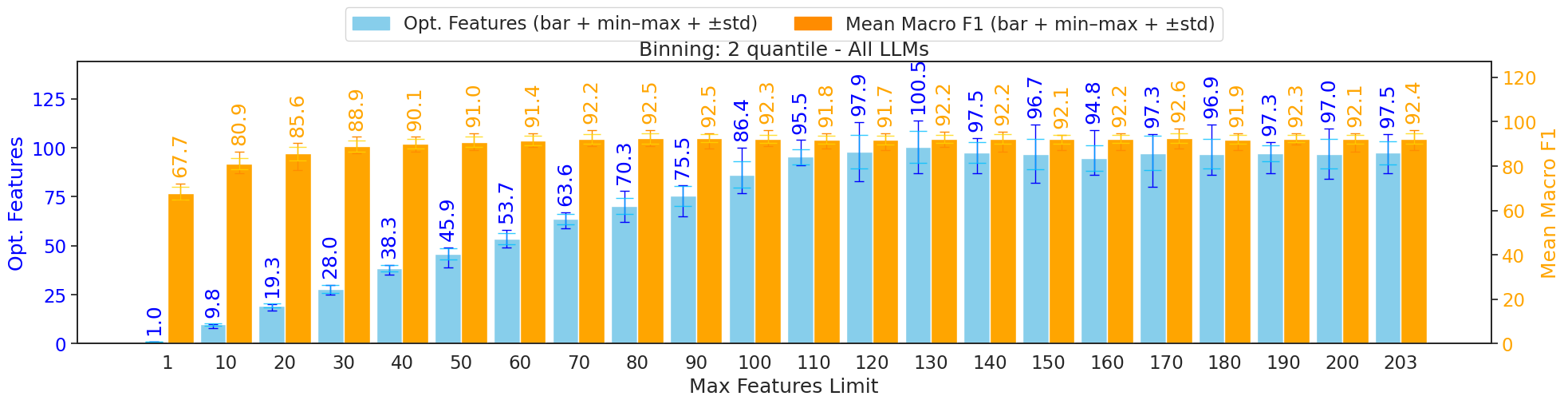}
    \includegraphics[width=1\linewidth]{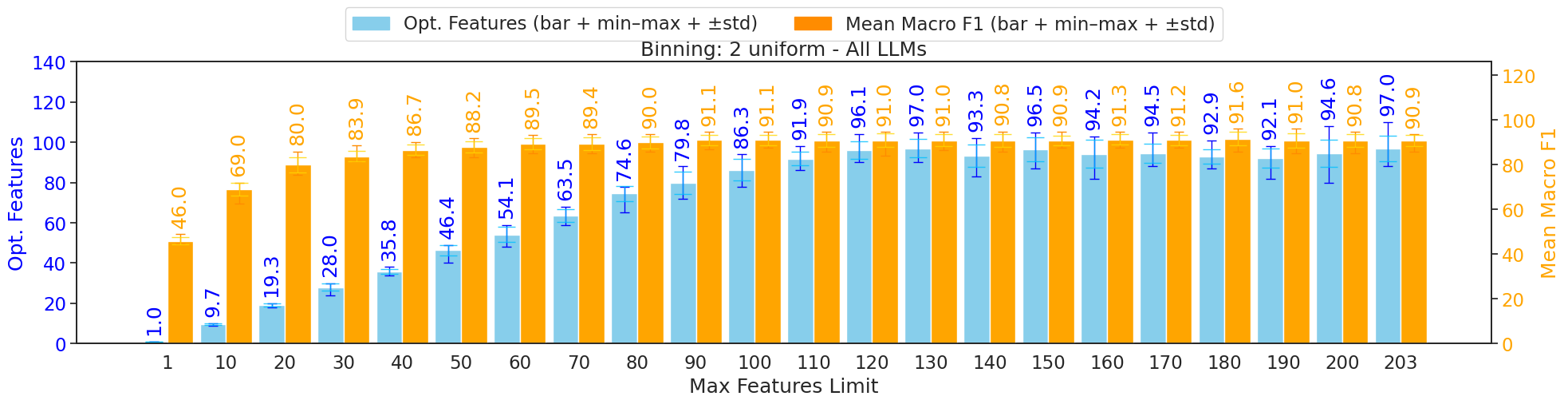}
    \includegraphics[width=1\linewidth]{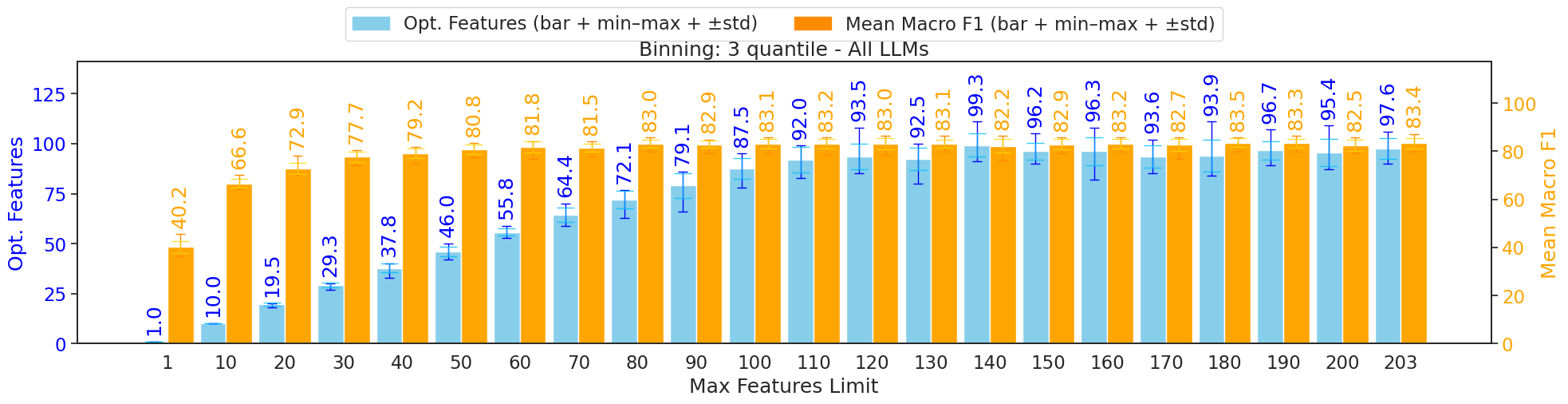}
    \includegraphics[width=1\linewidth]{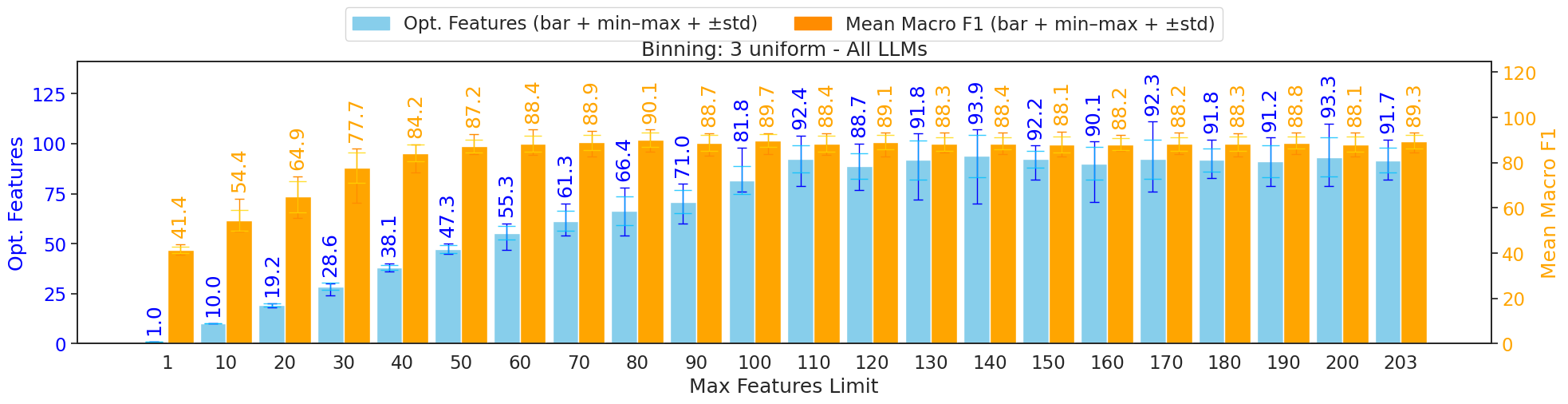}
    \caption{%
    Results of the classifier performance, optimised using an evolutionary search using the features of all LLMs as input.
    The y-axis shows the optimised Macro F1 score averaged over the 10 folds as well as the average number of selected features.
    The x-axis displays the upper number of features that the search algorithm was allowed to select.
    }
    \label{fig:results:evolsearch:all_llms}
\end{figure}

\begin{figure}[ht!]
    \centering
    \includegraphics[width=1\linewidth]{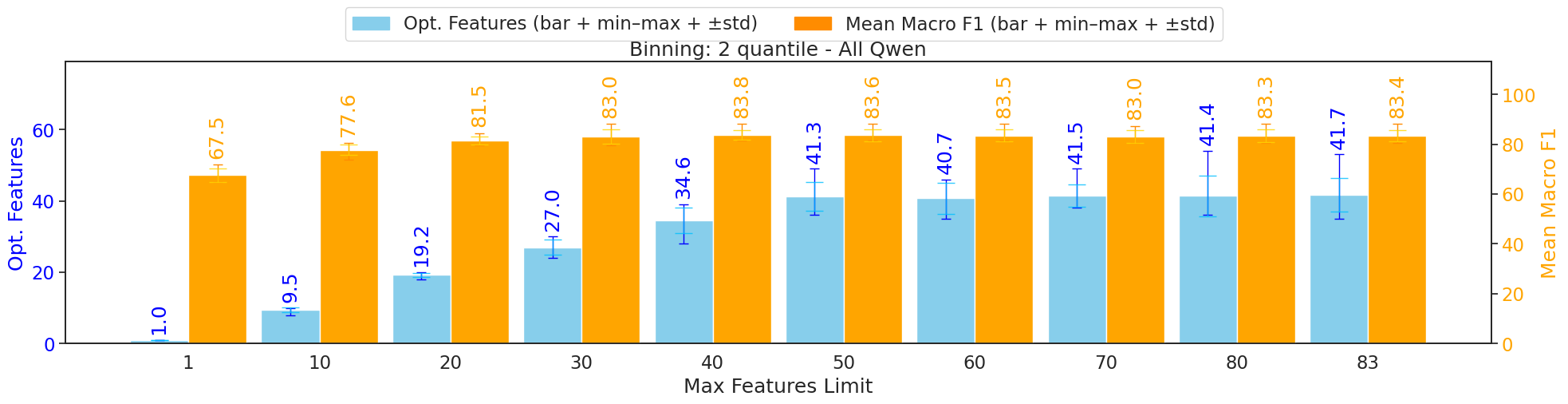}
    \includegraphics[width=1\linewidth]{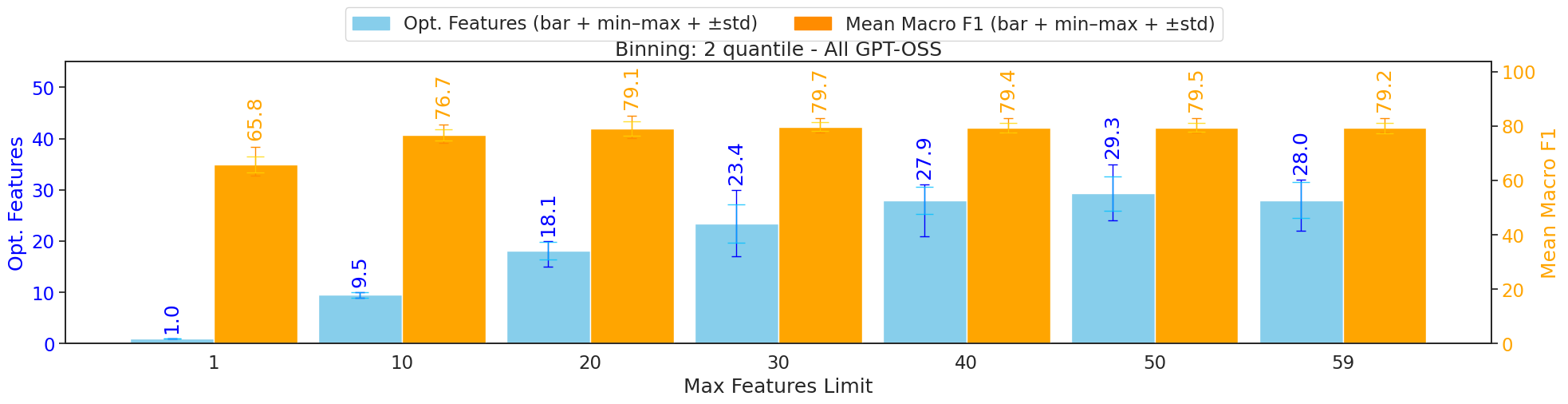}
    \includegraphics[width=1\linewidth]{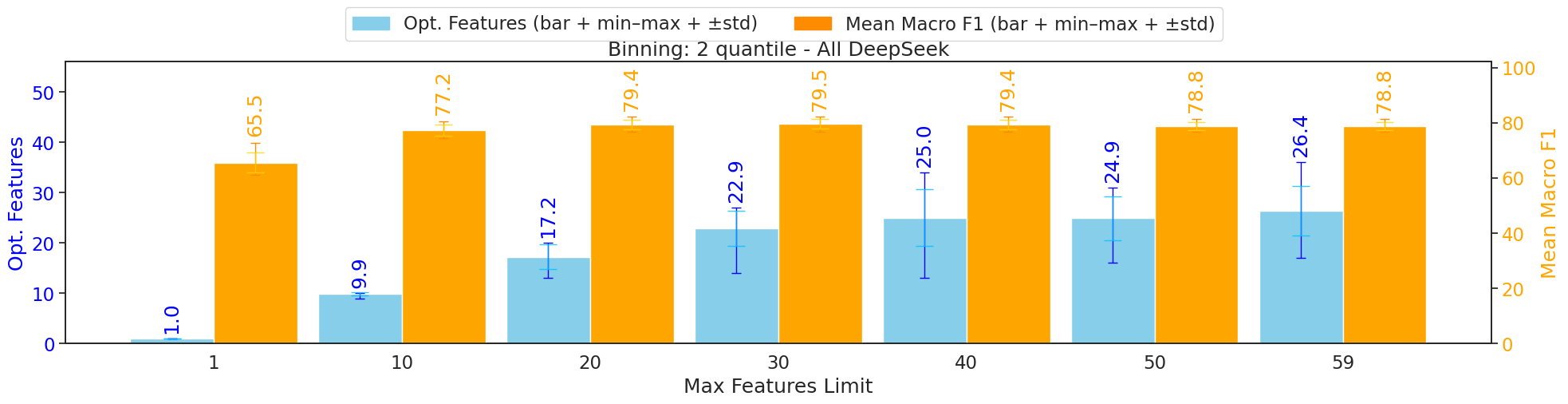}
    \caption{%
    Results of the \emph{2-quantile} binning variant classification optimised using an evolutionary search for the LLM-family groups.
    The images show the optimised mean Macro F1 score over 10 folds as well as the average number of selected features.
    }
    \label{fig:results:evolsearch:2_quantile}
\end{figure}

\begin{figure}[ht!]
    \centering
    \includegraphics[width=1\linewidth]{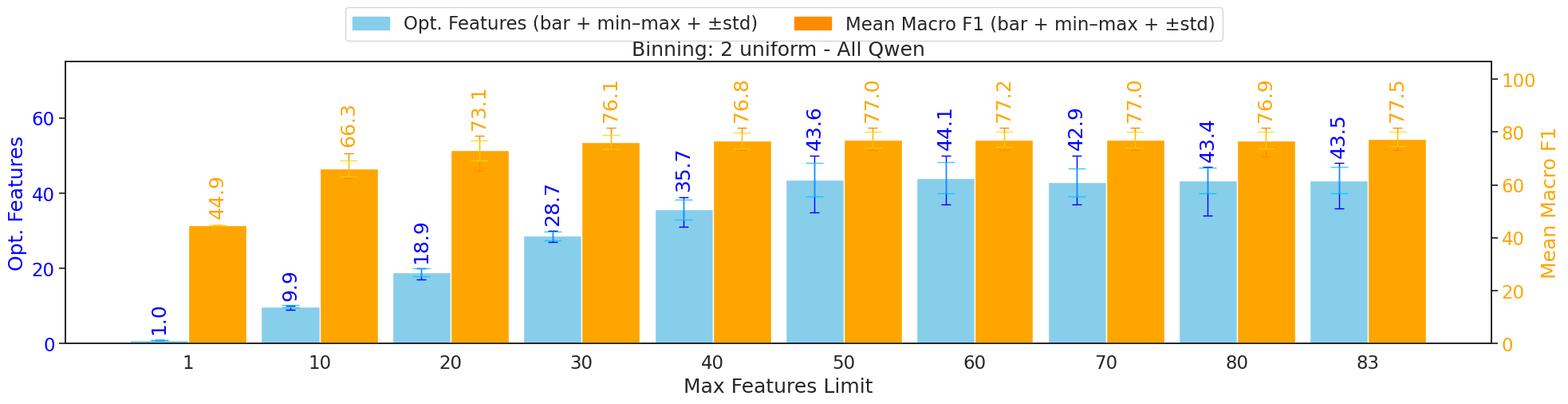}
    \includegraphics[width=1\linewidth]{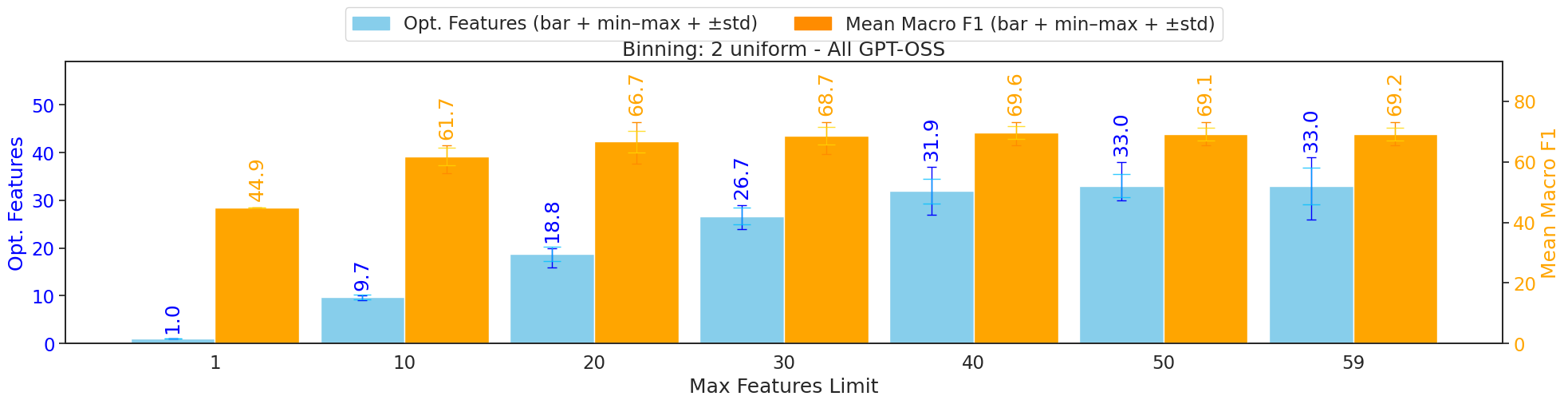}
    \includegraphics[width=1\linewidth]{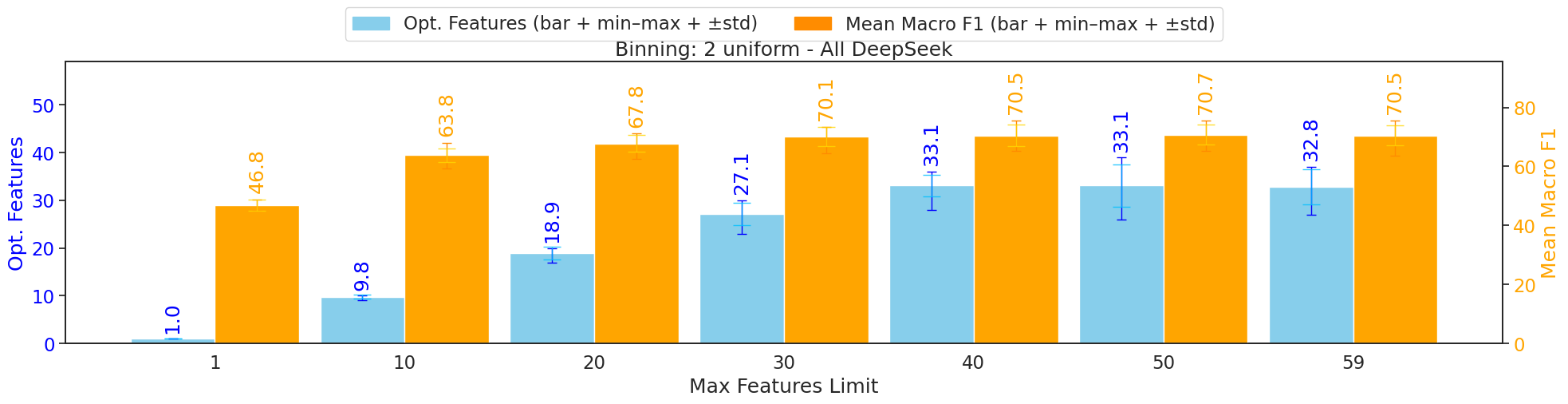}
    \caption{%
    Results of the \emph{2-uniform} binning variant classification optimised using an evolutionary search for the LLM-family groups.
    The images show the optimised mean Macro F1 score over 10 folds as well as the average number of selected features.
    }
    \label{fig:results:evolsearch:2_uniform}
\end{figure}

\begin{figure}[!htbp]
    \centering
    \includegraphics[width=1\linewidth]{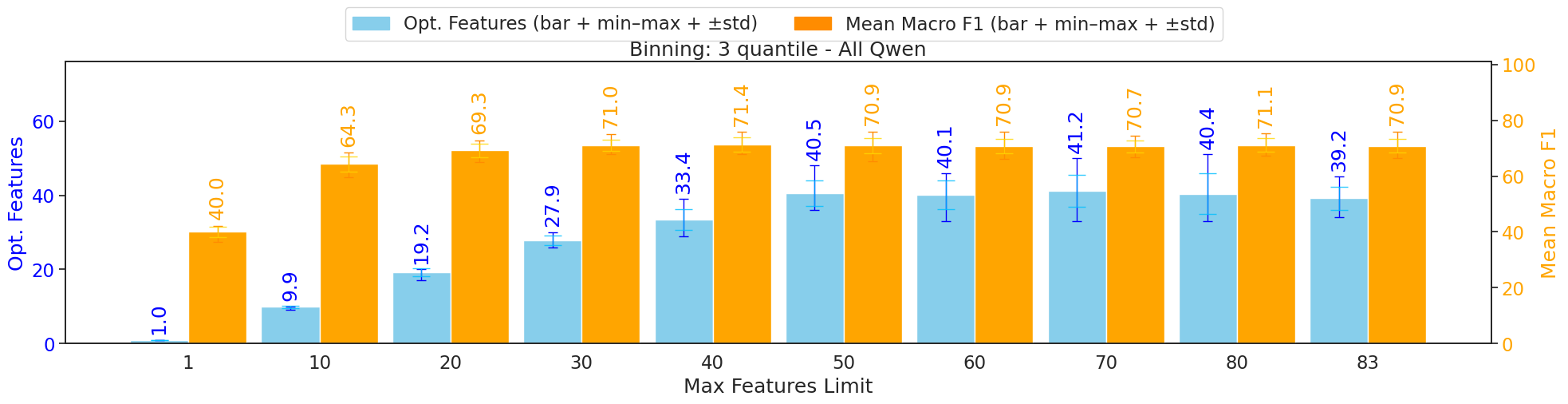}
    \includegraphics[width=1\linewidth]{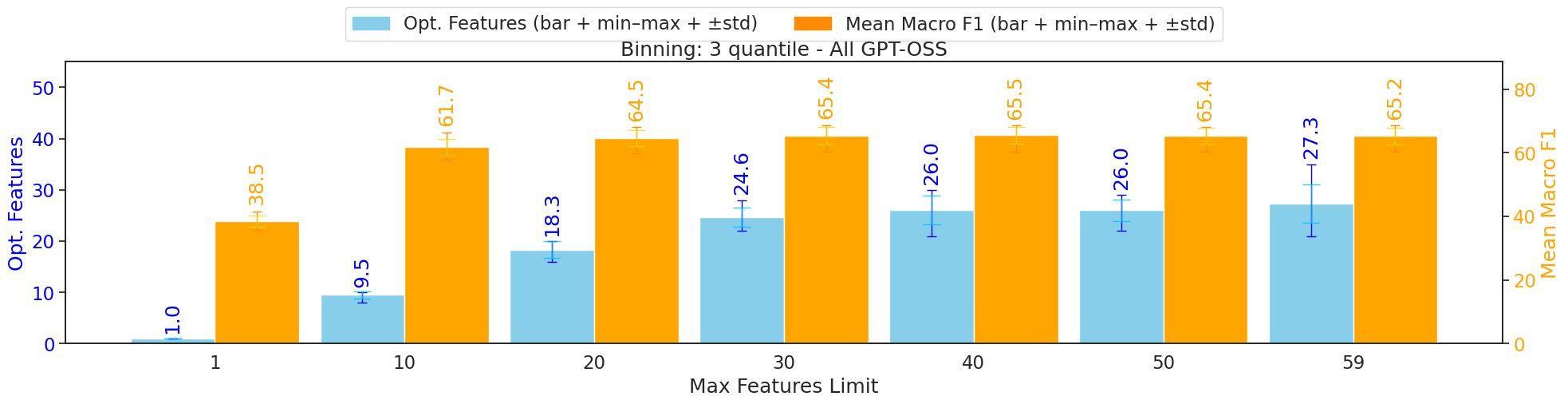}
    \includegraphics[width=1\linewidth]{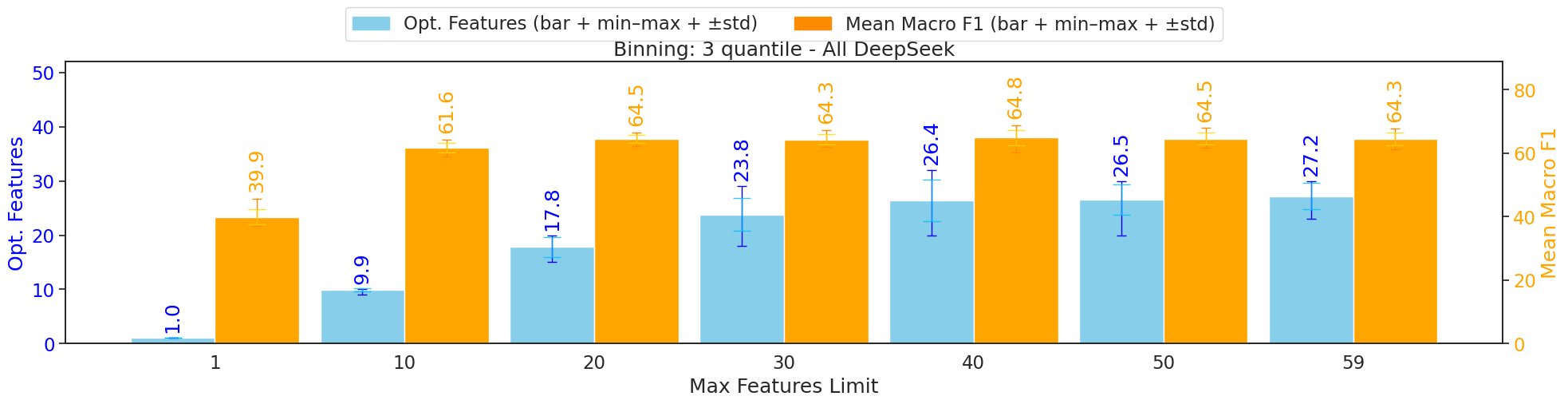}
    \caption{%
    Results of the \emph{3-quantile} binning variant classification optimised using an evolutionary search for the LLM-family groups.
    The images show the optimised mean Macro F1 score over 10 folds as well as the average number of selected features.
    }
    \label{fig:results:evolsearch:3_quantile}
\end{figure}

\begin{figure}[ht!]
    \centering
    \includegraphics[width=1\linewidth]{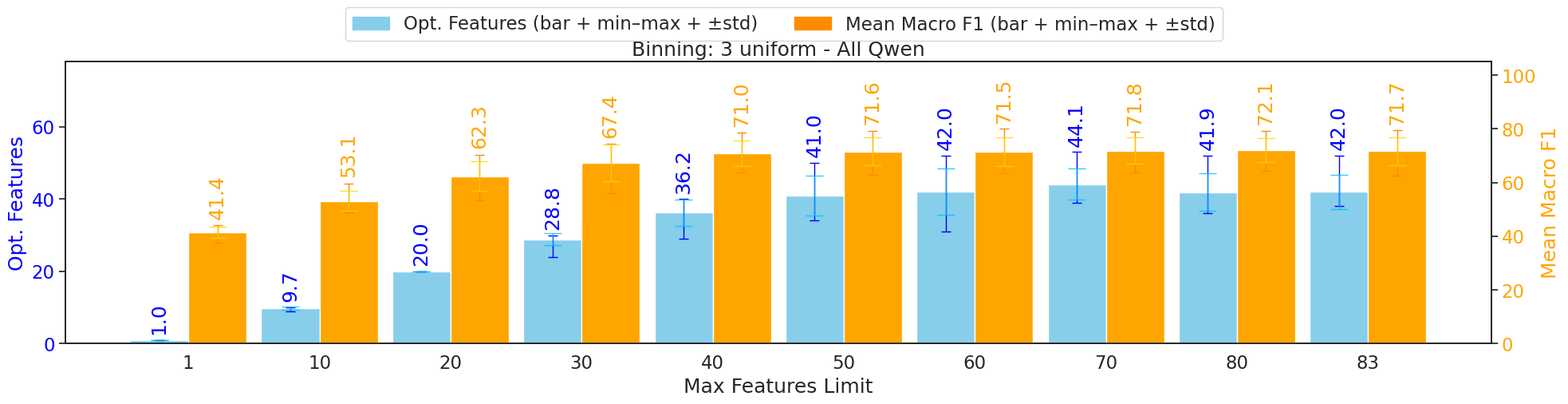}
    \includegraphics[width=1\linewidth]{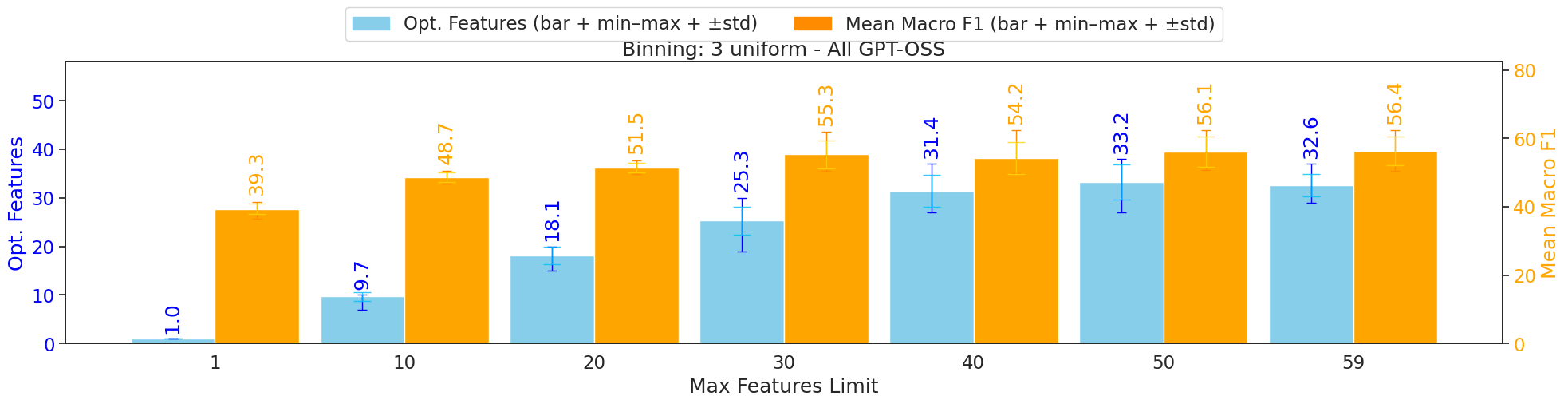}
    \includegraphics[width=1\linewidth]{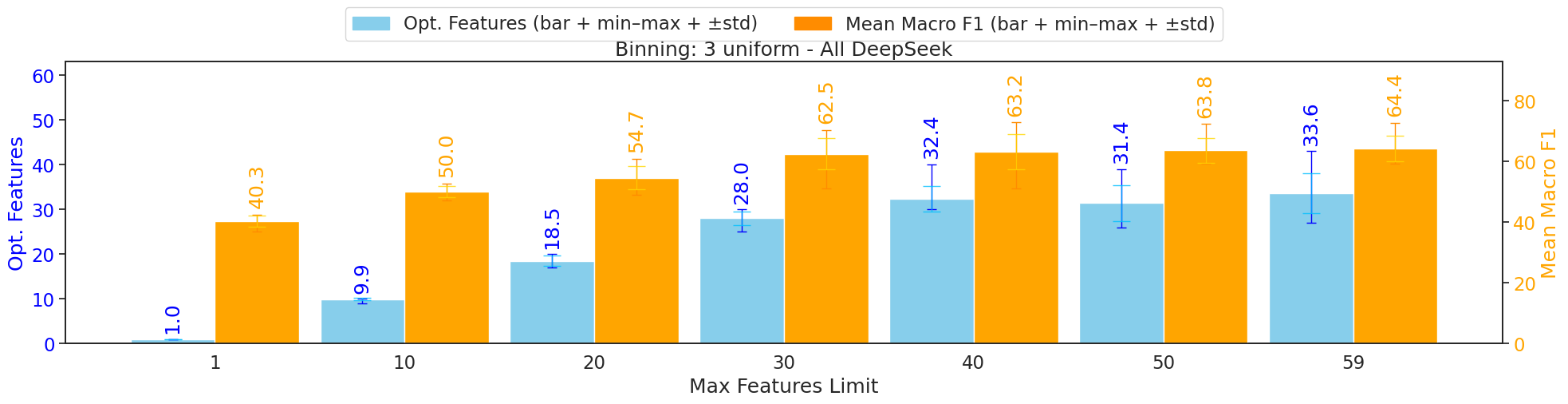}
    \caption{%
    Results of the \emph{3-uniform} binning variant classification optimised using an evolutionary search for the LLM-family groups.
    The images show the optimised mean Macro F1 score over 10 folds as well as the average number of selected features.
    }
    \label{fig:results:evolsearch:3_uniform}
\end{figure}

\begin{figure}[ht!]
    \centering
    \includegraphics[width=1\linewidth]{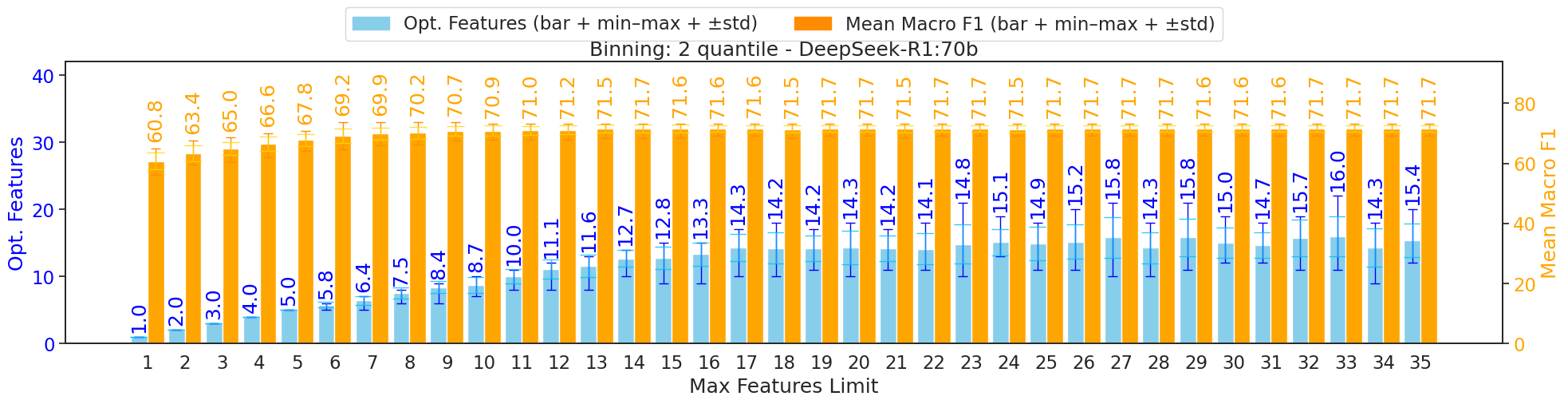}
    \includegraphics[width=1\linewidth]{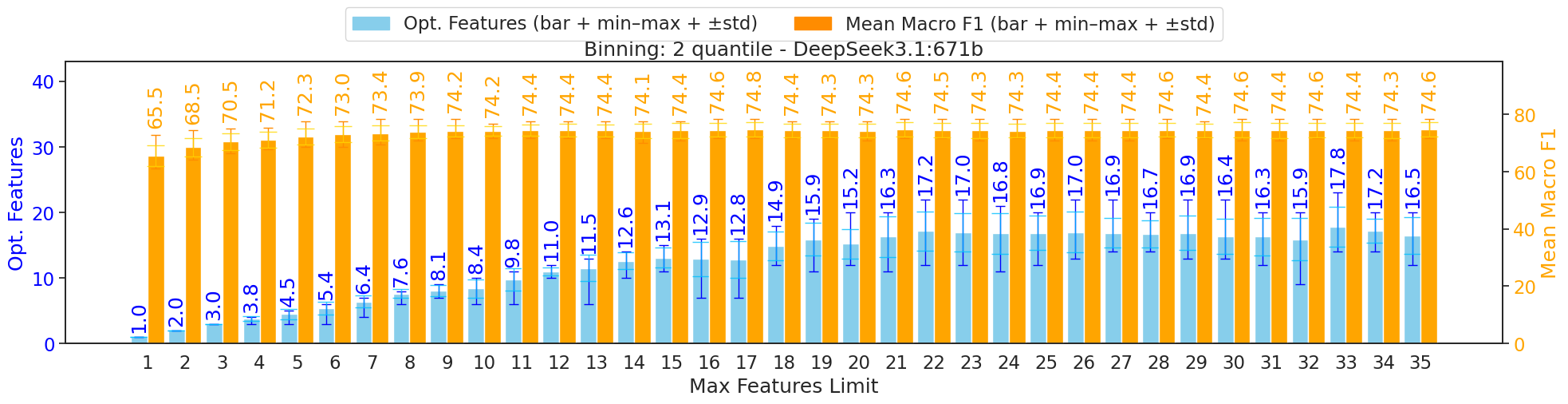}
    \includegraphics[width=1\linewidth]{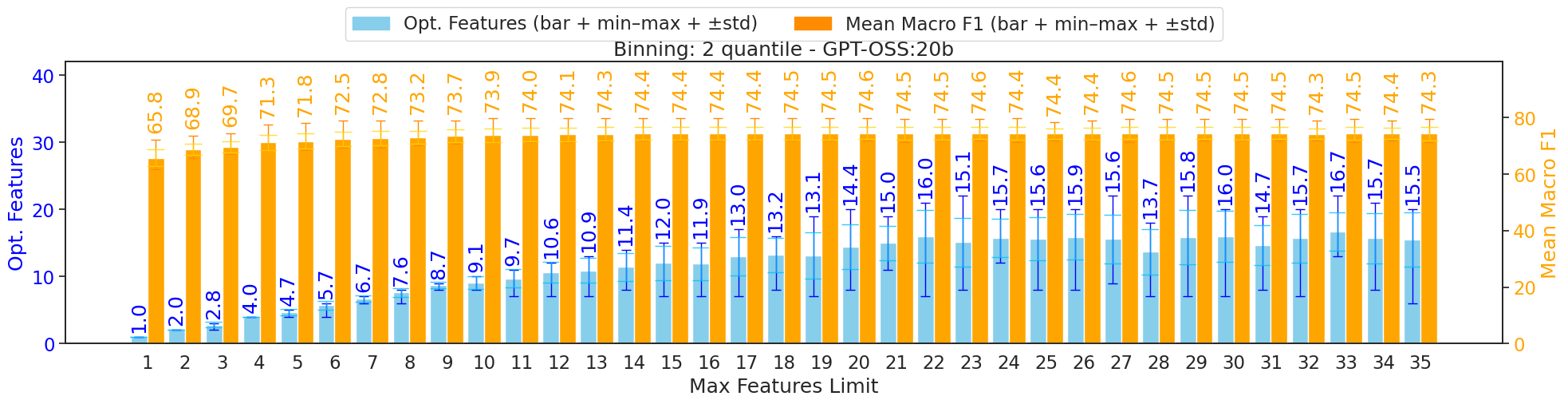}
    \includegraphics[width=1\linewidth]{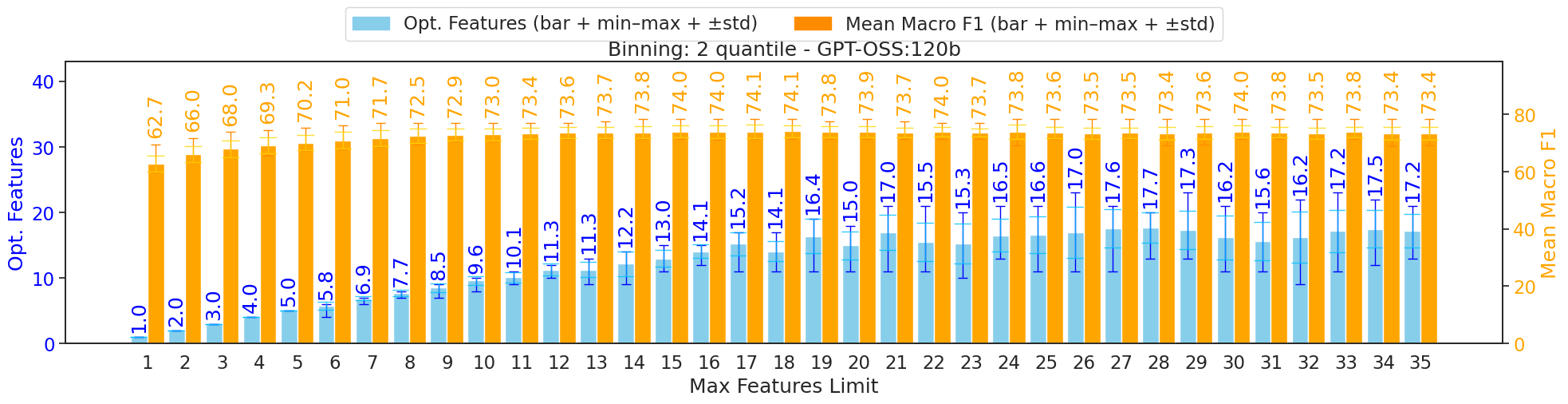}
    \caption{%
    Results of the \emph{2-quantile} binning variant classification optimised using an evolutionary search for the 2 DeepSeek- and the 2 GPT-OSS-LLMs.
    The images show the optimised mean Macro F1 score over 10 folds as well as the average number of selected features.
    }
    \label{fig:results:evolsearch:2_quantile_ds_gpt}
\end{figure}
\begin{figure}[ht!]
    \centering
    \includegraphics[width=1\linewidth]{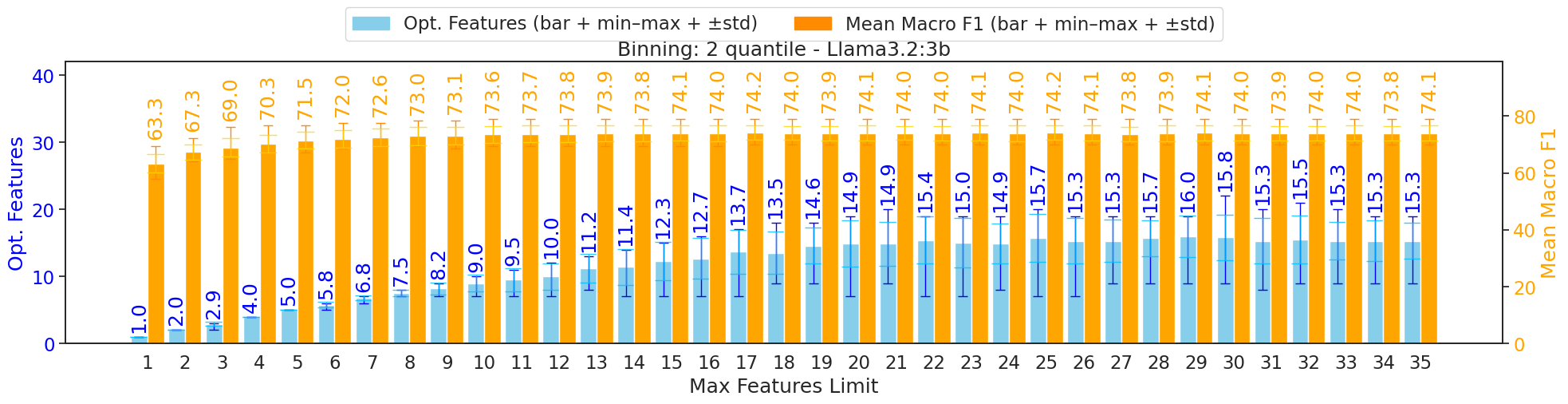}
    \includegraphics[width=1\linewidth]{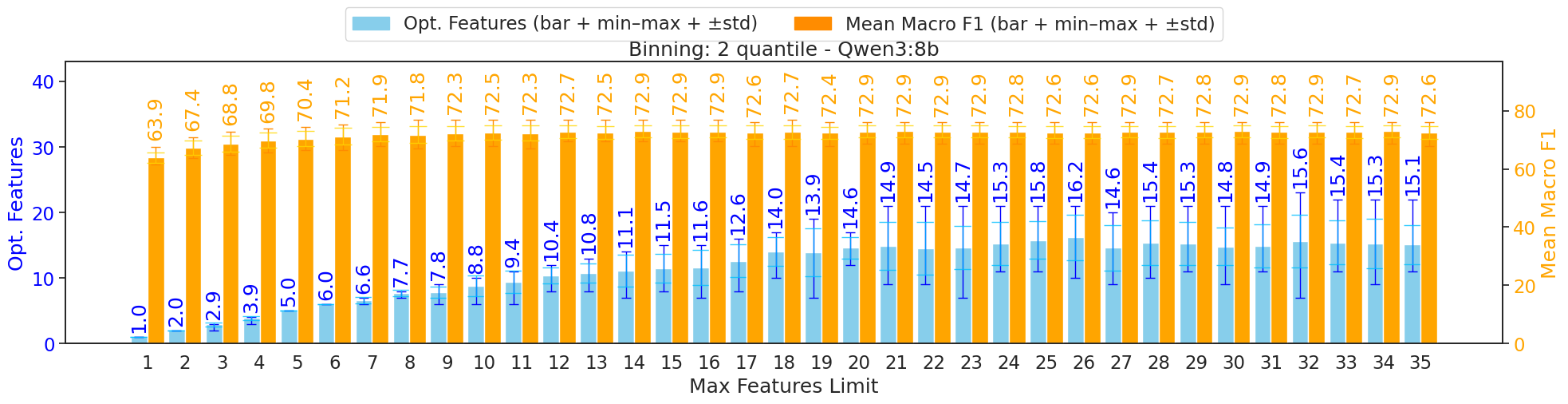}
    \includegraphics[width=1\linewidth]{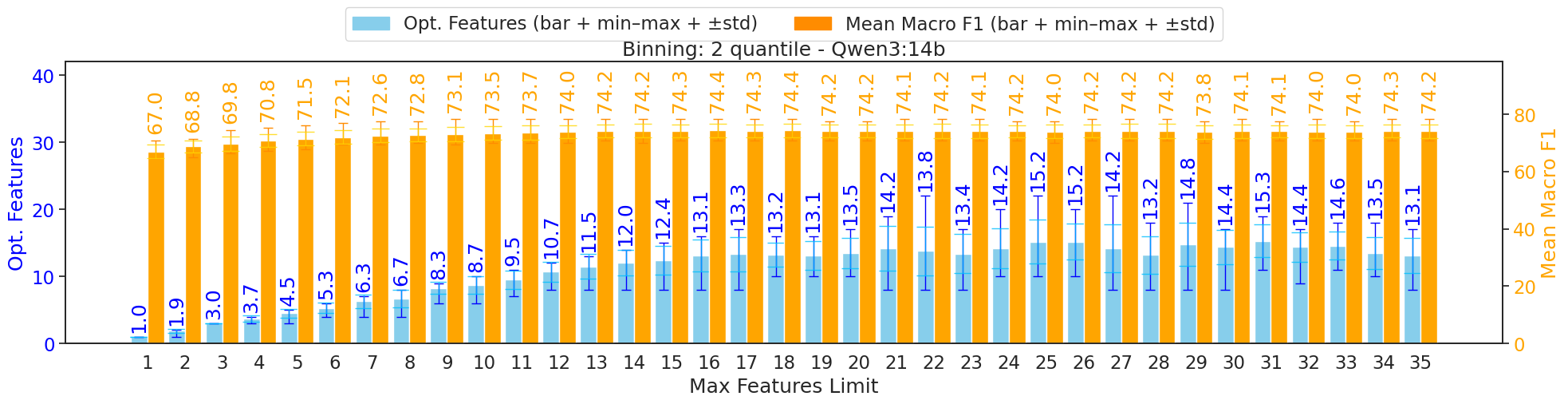}
    \includegraphics[width=1\linewidth]{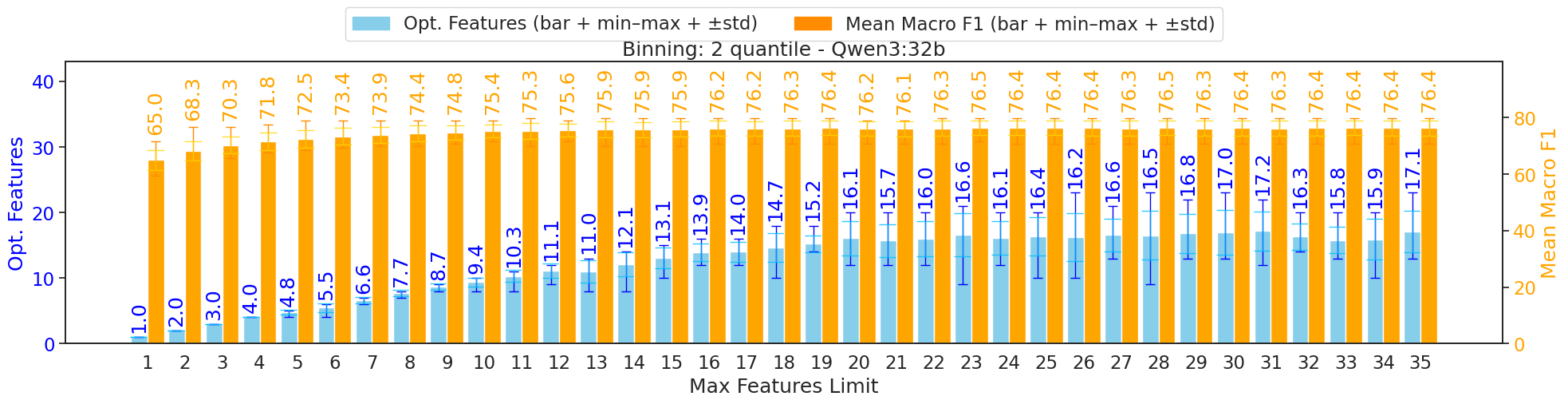}
    \caption{%
    Results of the \emph{2-quantile} binning variant classification optimised using an evolutionary search for the 3 Qwen-LLMs and Llama.
    The images show the optimised mean Macro F1 score over 10 folds as well as the average number of selected features.
    }
    \label{fig:results:evolsearch:2_quantile_2_llama_qwen}
\end{figure}

We start with experimenting with FG1 (\autoref{sec:methods}).
Our results show that a linear SVM trained on AES outputs can predict the performance classes of student responses.
We report the average Macro F1 for each of the 10 splits of the evolutionary search in \autoref{fig:results:evolsearch:all_llms}.
When trained on the combined outputs of all LLM-based AES, the classifier reaches Macro F1 of 92\% (quantile binning) and 91\% (uniform) for the binary classification task.
For the three-class task, it achieves Macro F1 of 83\% (quantile) and 89\% (uniform).
Across all settings, about half of the 203 features are sufficient to achieve this result. Even with 20-30 features, the classifier still achieves comparable scores.
A simple linear SVM can use small AES-based feature vectors to achieve nearly perfect classification ($>$ 90\%), so H1 is not falsified: despite their general information and knowledge bases, LLM-based AES produce features that almost fully reconstruct expert-derived performance classes. This result is achieved without extensive hyperparameter tuning.

We show detailed results for FG2 and FG3 (again based on linear SVMs):
Performance by LLM family (FG3) is shown in Figures~\ref{fig:results:evolsearch:2_quantile} to~\ref{fig:results:evolsearch:3_uniform} and performance for each of the eight LLMs (FG2) is reported in Figures~\ref{fig:results:evolsearch:2_quantile_ds_gpt} and~\ref{fig:results:evolsearch:2_quantile_2_llama_qwen}.
We observe a lower Macro F1 in these settings, but the differences between LLM families remain small. GPT-OSS and DeepSeek perform about 15\% worse than Qwen in the three-class uniform variant and about 5\%-10\% worse in the remaining three variants.
The Qwen LLM family performs best across all variants, both when using all 83 Qwen features and when restricted to 60 features, outperforming GPT-OSS and DeepSeek (59 features each).
This suggests that feature count is not the primary factor driving these results.
In the eight single-model settings (35 features each), the classifier achieves Macro F1 of 72\%-76\% (quantile) and 49\%-63\% (uniform) for the binary task, and 56\%-61\% (quantile) and 47\%-56\% (uniform) for the three-class task.
\begin{figure}[ht!]
    \centering
    \includegraphics[width=1\linewidth]{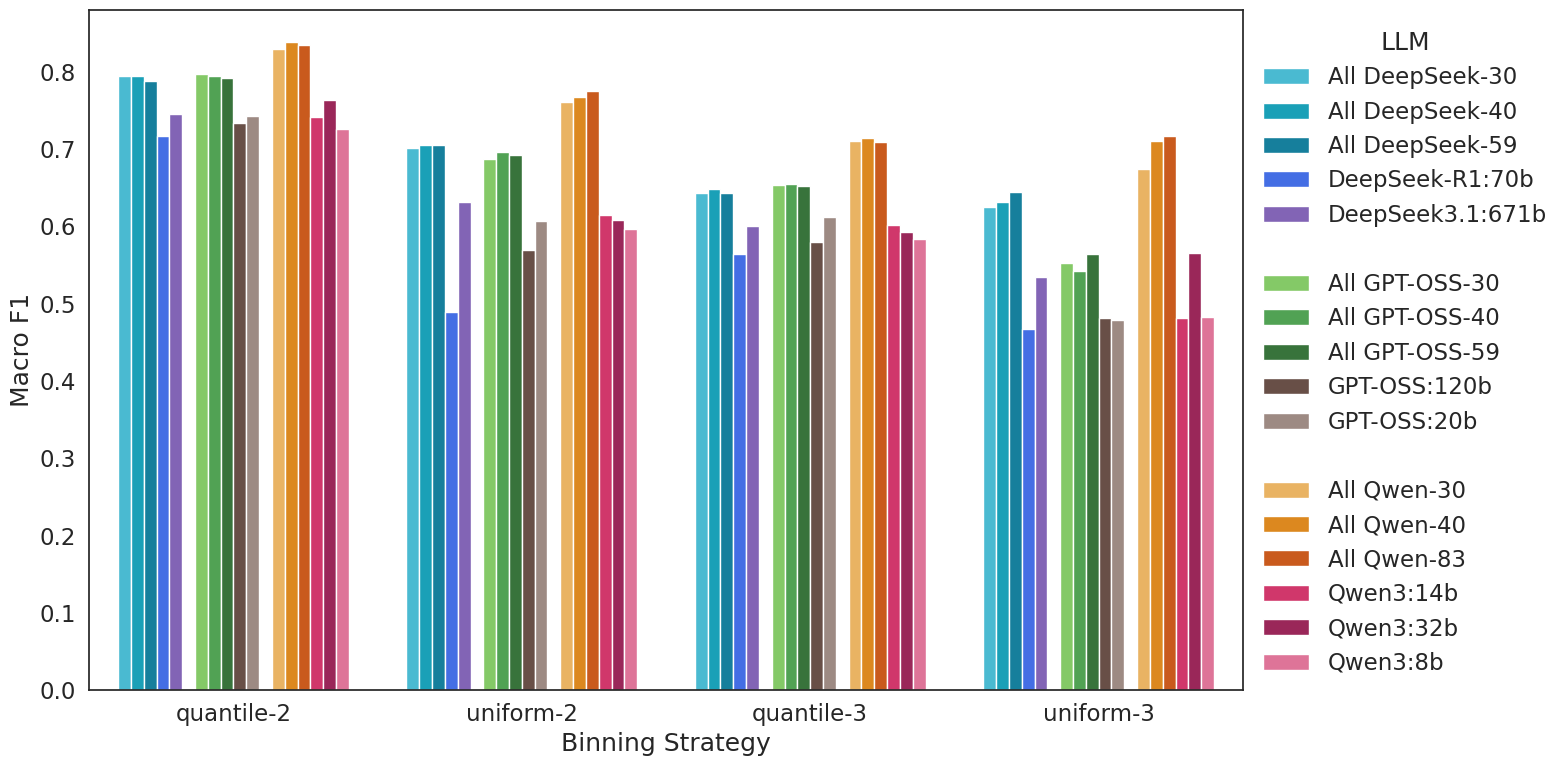}
    \caption{
    Barplot showing the mean Macro F1 scores of the classification performance after optimization, comparing the results of the feature selection by single LLM vs. LLM-families.
    }
    \label{fig:llm_family_improvement}
\end{figure}
Combining the features of multiple models within the same LLM family improves performance over the best individual model across all binnings (see \autoref{fig:llm_family_improvement}). This improvement is not merely due to having more features, as gains persist even when training is limited to 30-40 features.

\begin{figure}[!htbp]
    \centering
    \includegraphics[width=1\linewidth]{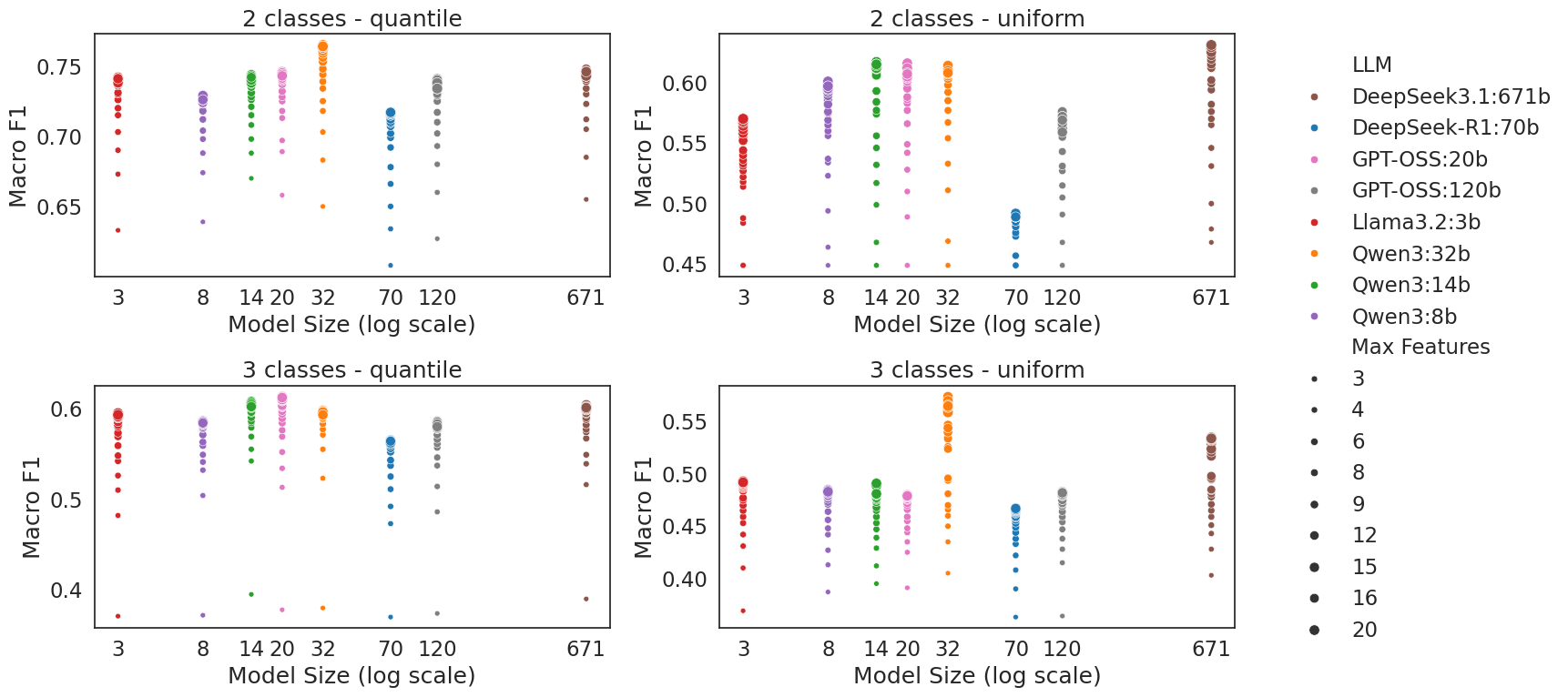}
    \caption{
    Scatterplot of LLM parameter size (in billions of parameters) versus mean Macro F1 after evolutionary search. Point size indicates the maximum number of features permitted in the search.
    }
    \label{fig:llm_parameters_influence}
\end{figure}

\autoref{fig:llm_parameters_influence} shows the effect of LLM parameter size on Macro F1 after evolutionary search, revealing no relationship between model size and performance across all class configurations: LLM size is not a significant factor in our classification scenario.

\begin{table}[!htbp]
    \centering
    \begin{tabular}{l|r|r|r|r}
        \toprule
       Task & 2-quantile & 2-uniform & 3-quantile & 3-uniform \\
       \midrule
       Middle Ear & 0.90 & 0.95 & 0.80 & 0.83 \\
       Respiration & 0.92 & 0.90 & 0.82 & 0.87 \\
       Eye & 0.90 & 0.84 & 0.81 & 0.85 \\
       Circulation & 0.95 & 0.92 & 0.87 & 0.93 \\
         \bottomrule
    \end{tabular}
    \caption{Mean Macro F1 scores of the classifier performance after optimization using evolutionary search with feature inputs from all LLMs, broken down by CORE assessment task (first column).}
    \label{tab:f1_scores_per_task}
\end{table}

\autoref{tab:f1_scores_per_task} reports Macro F1 for each task. Although trained on all medical tasks jointly, the classifier generalises well and predicts performance classes with high accuracy, indicating minimal task dependence and showing that AES features generalise across diverse medical tasks.

\subsection{Feature Importance Analysis}\label{sec:results:distribution_analysis}
To compare the class distributions of the individual features, we used a non-parametric approach, combining a rank-based effect size with a distribution-free significance test.
We focus on the binary quantile binning strategy for the performance classes in this analysis.
For each feature $f$, the difference between the two classes (\emph{high} and \emph{low}) was quantified using Cliff's delta $\delta_f$ \cite{Cliff:1993} and its statistical significance assessed with a two-sided Mann--Whitney $U$ test \cite{Mann:Whitney:1947}.
For samples $\{x_i\}_{i=1}^{m}$ from class \emph{high} and $\{y_j\}_{j=1}^{n}$ from class \emph{low}, Cliff's delta was computed as
\[
\delta_f = \frac{1}{mn}\sum_{i=1}^{m}\sum_{j=1}^{n} \big[\mathbb{I}(x_i>y_j)-\mathbb{I}(x_i<y_j)\big] = \Pr(X>Y)-\Pr(X<Y),
\]
with $\delta_f \in [-1,1]$.
Positive values of Cliff's delta indicate that the feature has a tendency to be larger in class \emph{low}, while negative values indicate larger values in class \emph{high}.
For the Mann--Whitney test, all $m+n$ observations were ranked jointly (averaging ranks in case of ties), allowing for unequal sample sizes between classes.
Let $R_x$ denote the sum of ranks of the $m$ observations from class \emph{high}. The test statistic was then computed as
\[
U = R_x - \frac{m(m+1)}{2},
\]
which equals the number of pairwise comparisons where a value from class \emph{high} exceeds one from class \emph{low}, plus half of the ties.
The corresponding two-sided $p$-value, $p_f$, was obtained under the null hypothesis of equal distributions, using either the exact or asymptotic null distribution depending on sample size and the presence of ties.
To control for multiple testing across all features, the Benjamini–Hochberg (BH) false discovery rate (FDR) procedure \cite{Benjamini:Hochberg:1995} was applied at level $q=0.05$.

\begin{figure}[!htbp]
    \centering
    \includegraphics[width=0.5\linewidth]{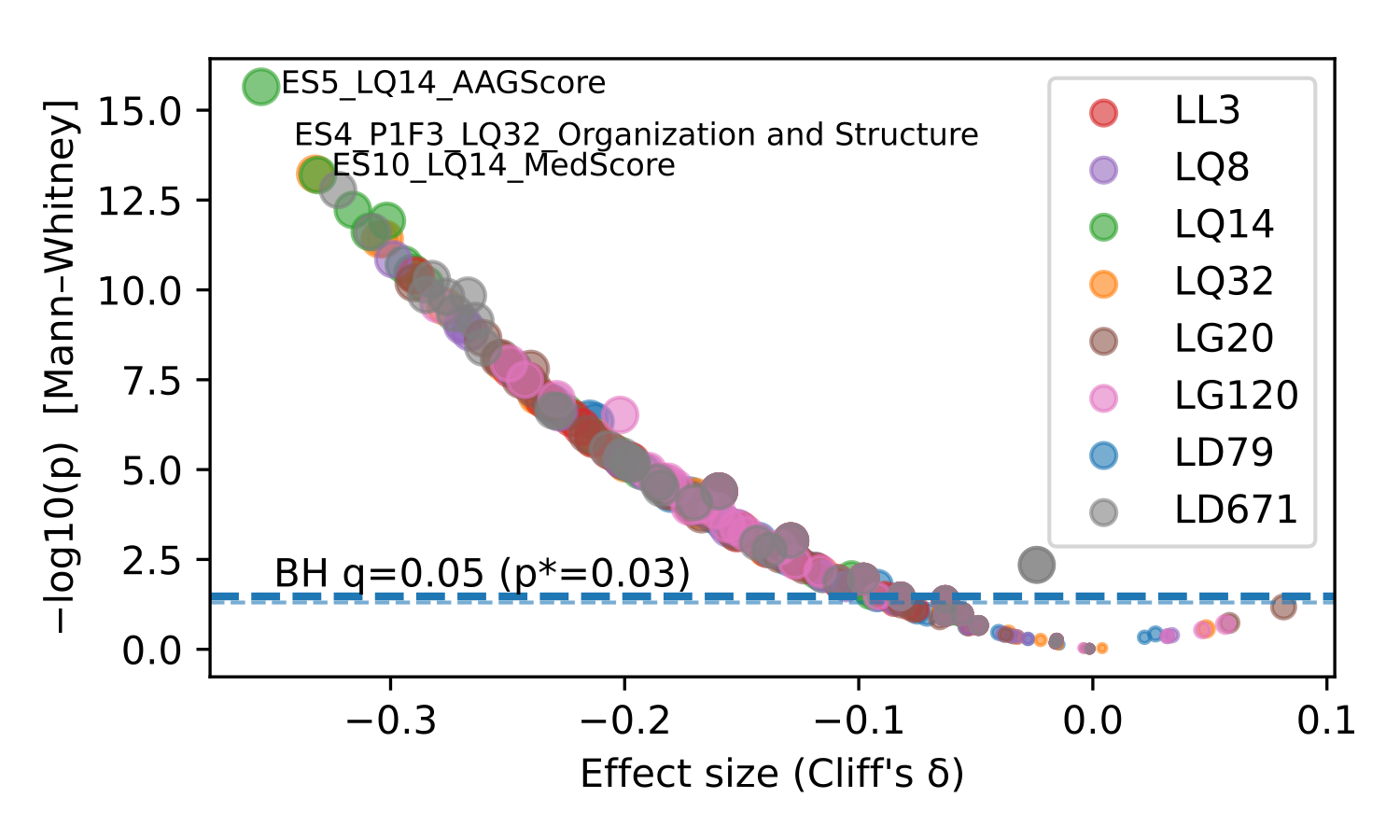}
    \caption{
    Volcano plot comparing distributions for all models.
    The x-axis shows Cliff's delta \((\delta\in[-1,1])\), where \(\delta>0\) favours class \emph{low} and \(\delta<0\) favours \emph{high}.
    The y-axis shows \(-\log_{10}\) two-sided Mann--Whitney \(p\)-values (higher indicates stronger evidence).
    The dashed line marks \(p=0.05\); the FDR line marks \(q=0.05\) \cite{Benjamini:Hochberg:1995}.
    }
    \label{fig:volcano_all}
\end{figure}

\begin{figure}[!htbp]
    \centering
    \includegraphics[width=0.49\linewidth]{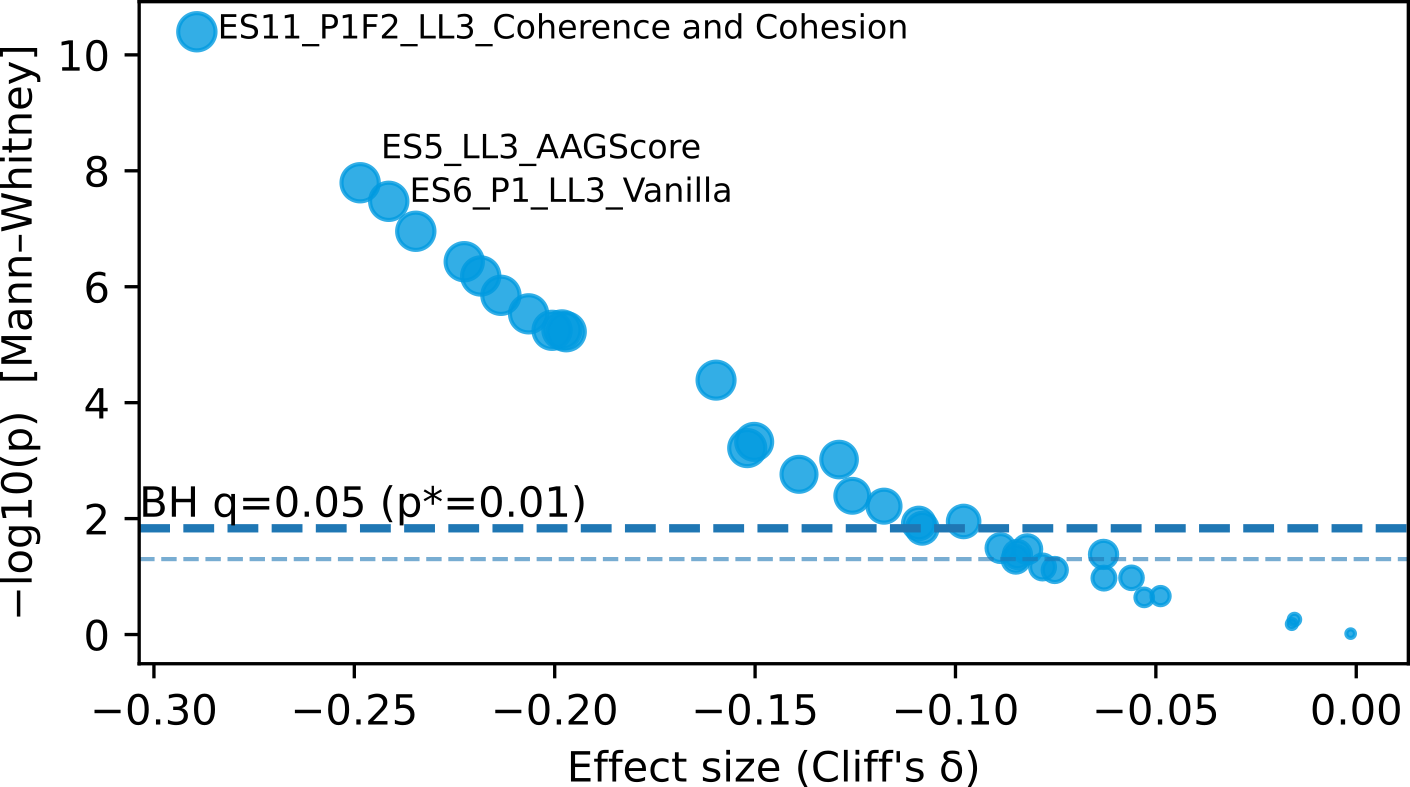}
    \includegraphics[width=0.49\linewidth]{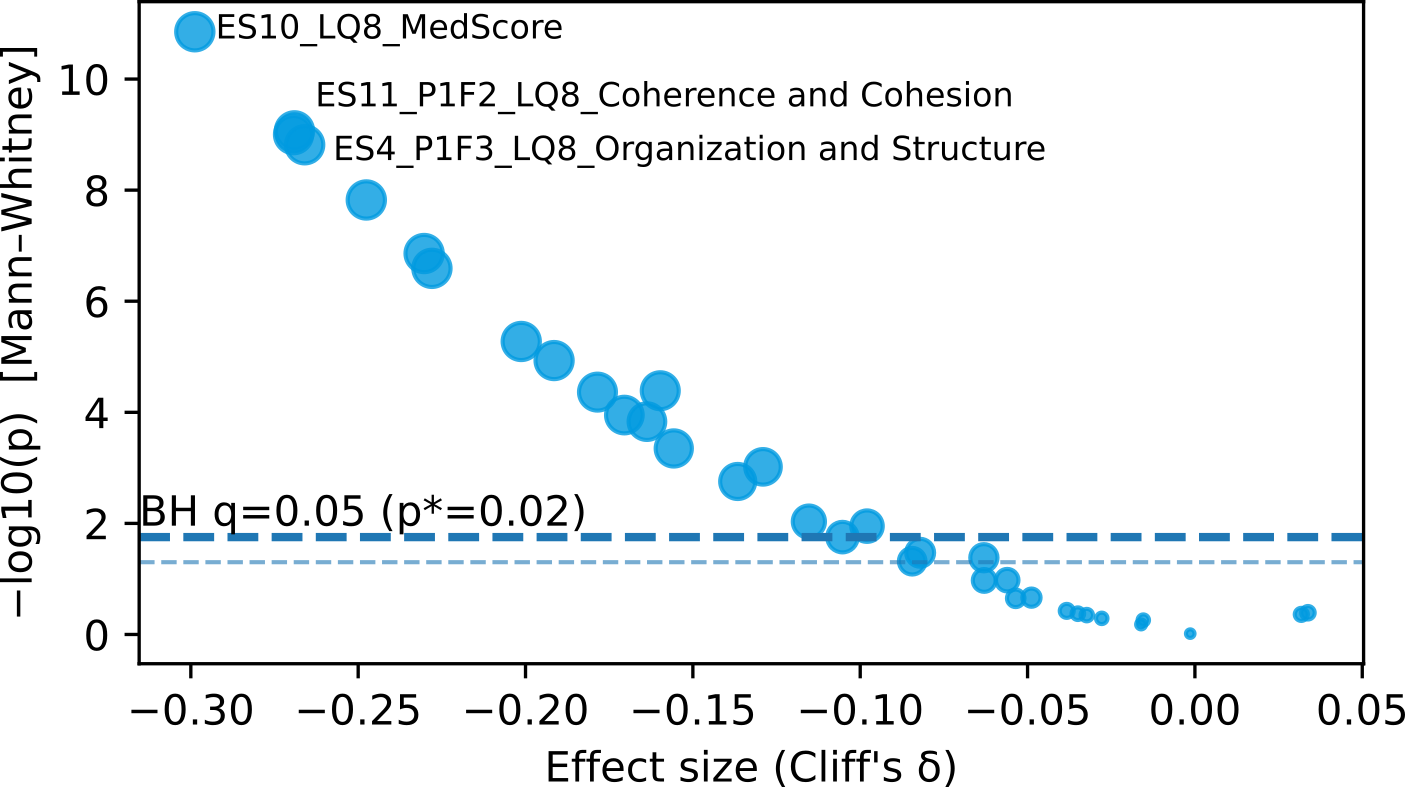}
    \includegraphics[width=0.49\linewidth]{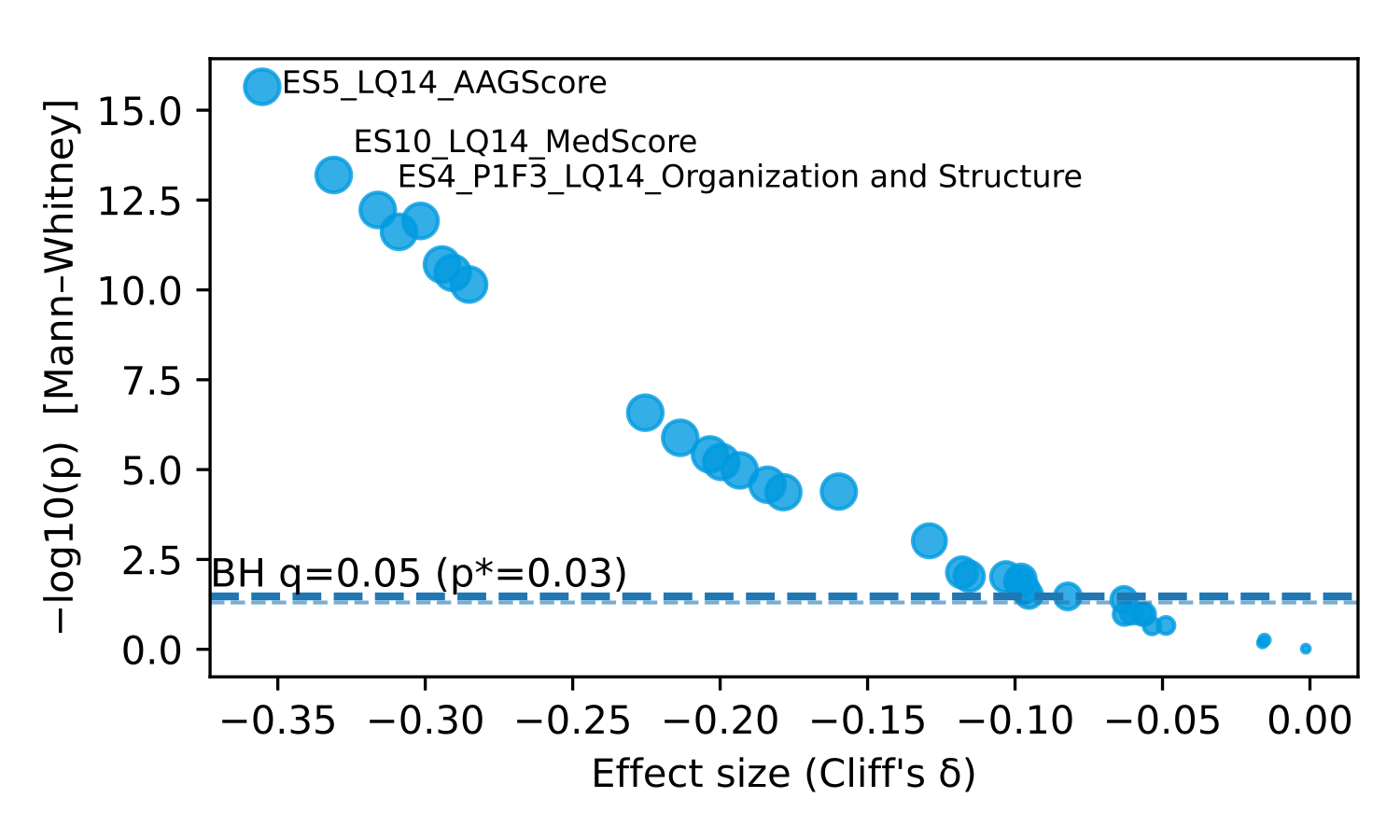}
    \includegraphics[width=0.49\linewidth]{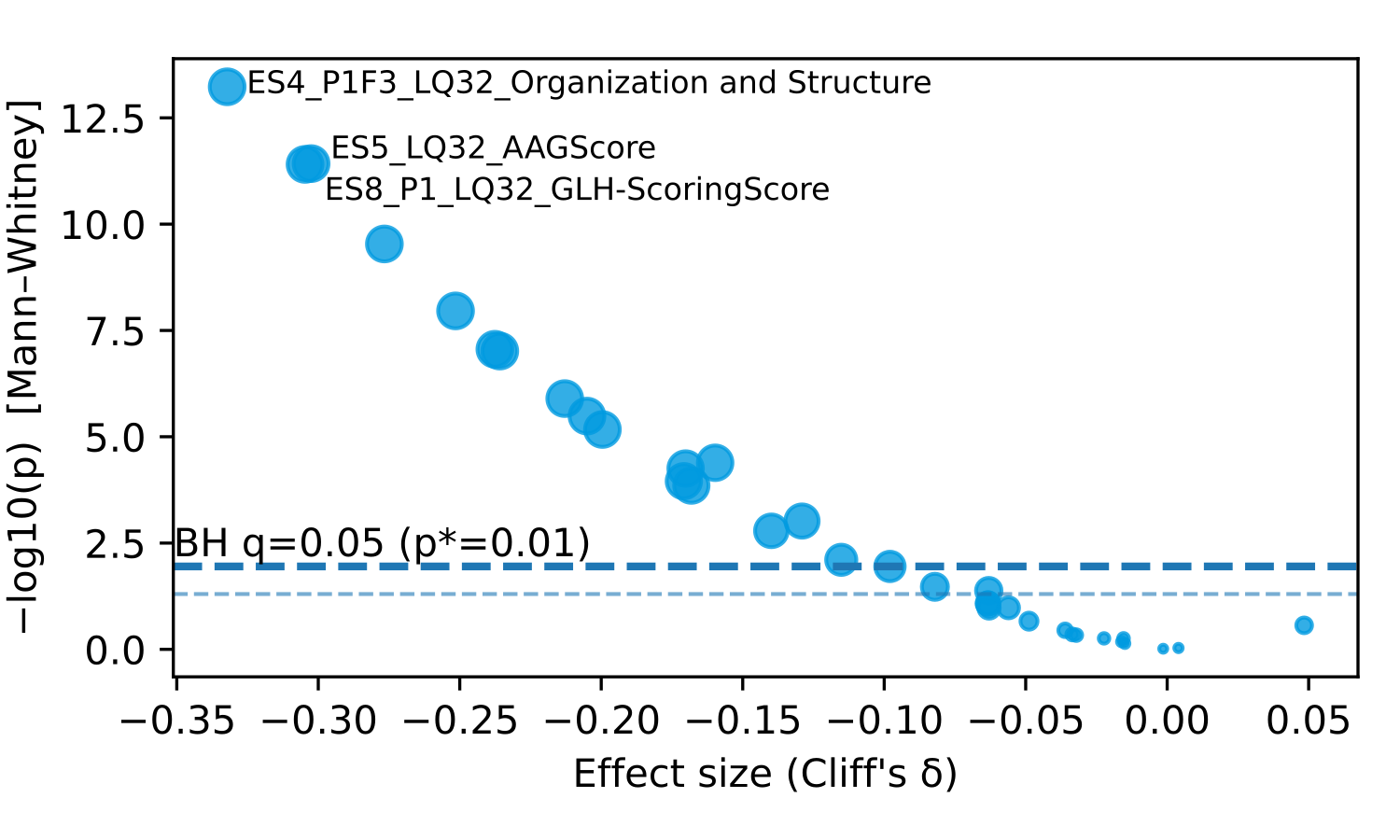}
    \caption{
    Volcano plot for LLAMA (top left) and Qwen models (same method as \autoref{fig:volcano_all}).
    }
    \label{fig:volcano_qwen}
\end{figure}

\begin{figure}[!htbp]
    \centering
    \includegraphics[width=0.49\linewidth]{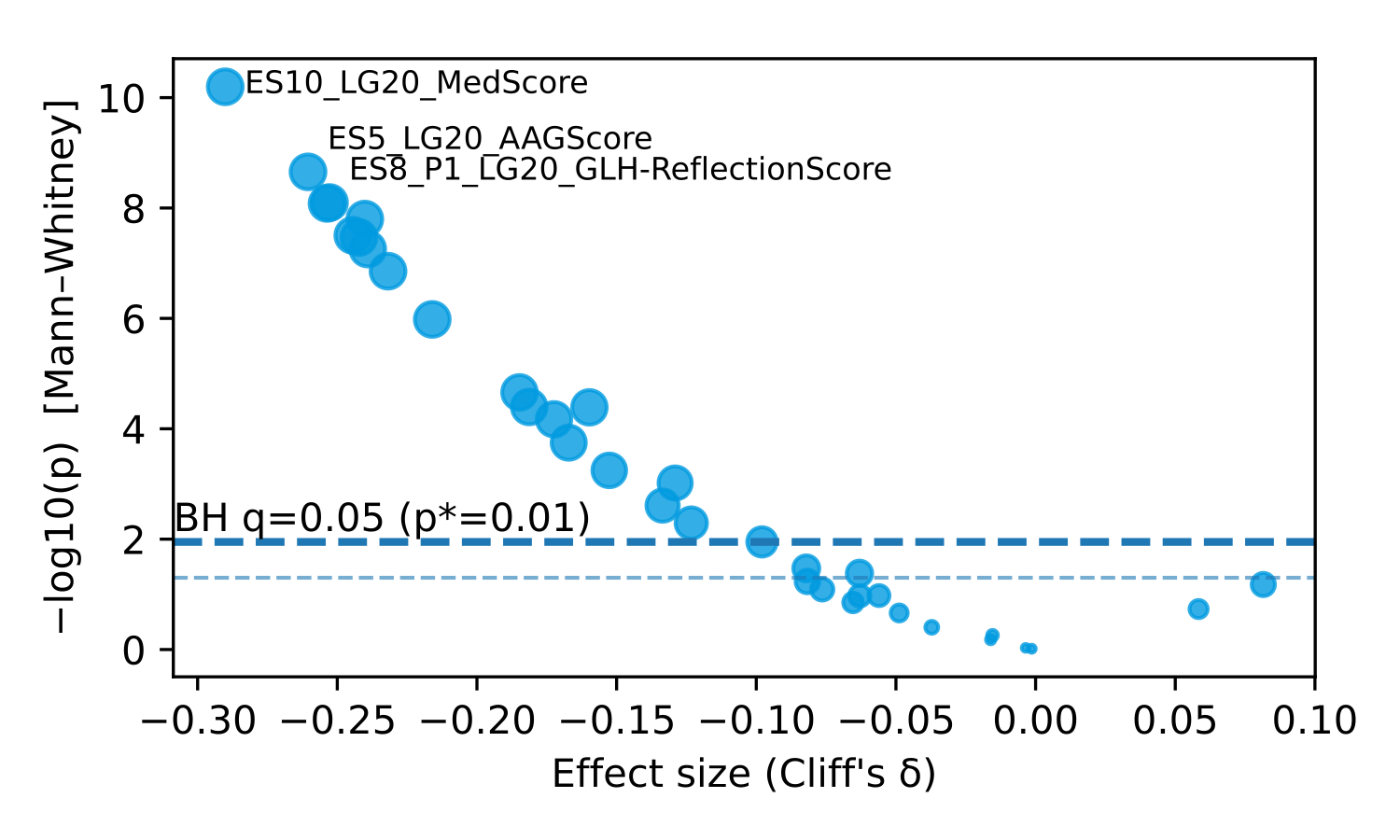}
    \includegraphics[width=0.49\linewidth]{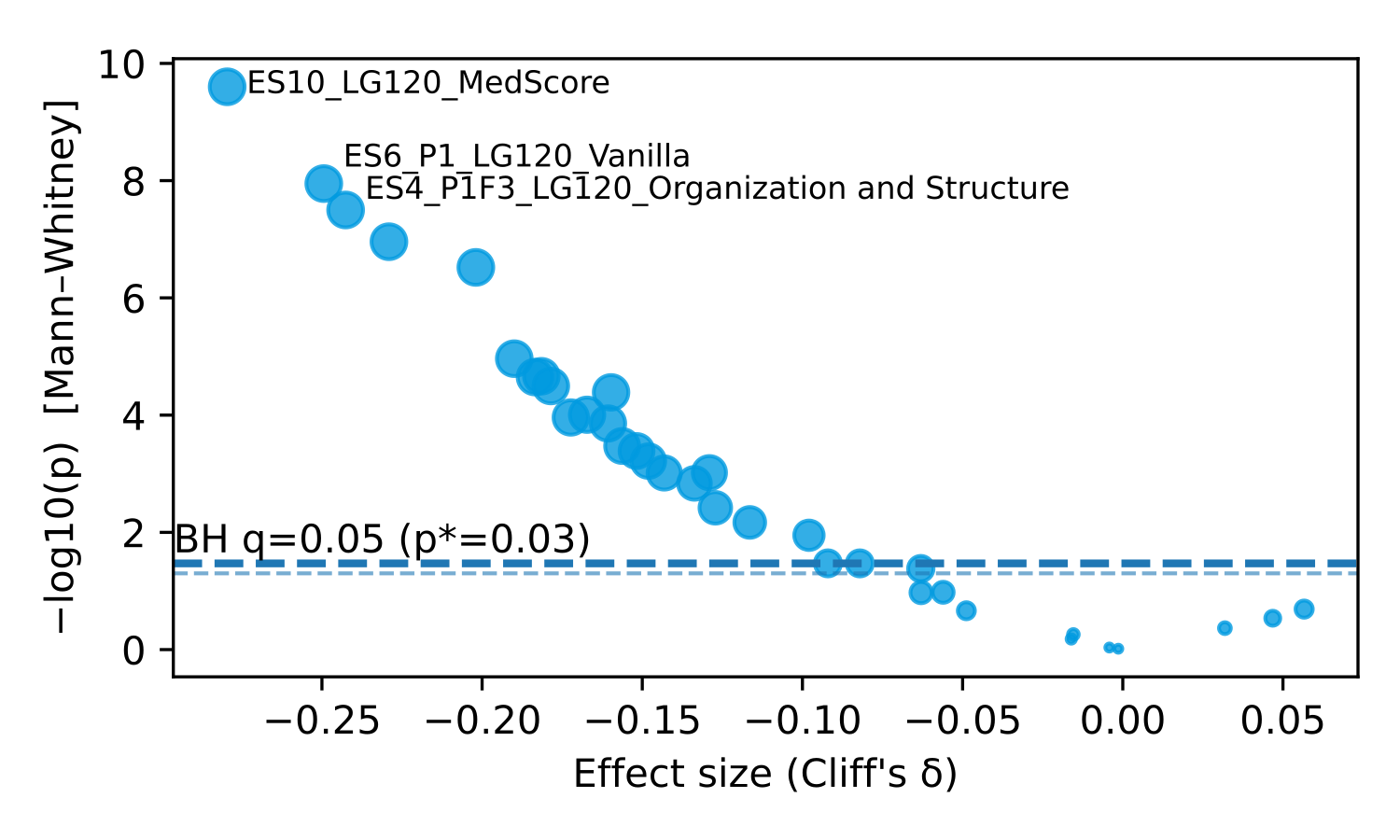}
    \caption{
    Volcano plot for GPT models (same method as \autoref{fig:volcano_all}).
    }
    \label{fig:volcano_qpt}
\end{figure}

\begin{figure}[!htbp]
    \centering
    \includegraphics[width=0.49\linewidth]{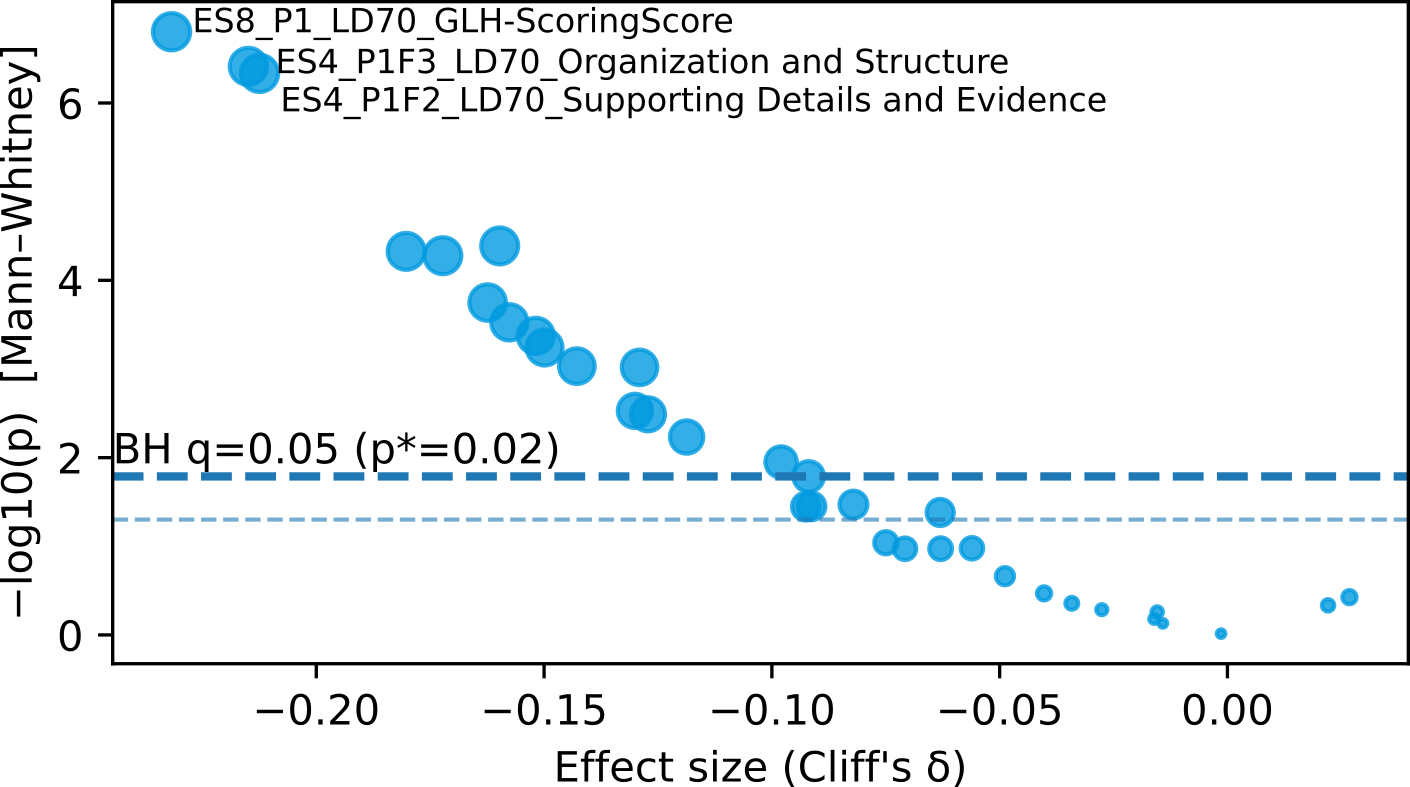}
    \includegraphics[width=0.49\linewidth]{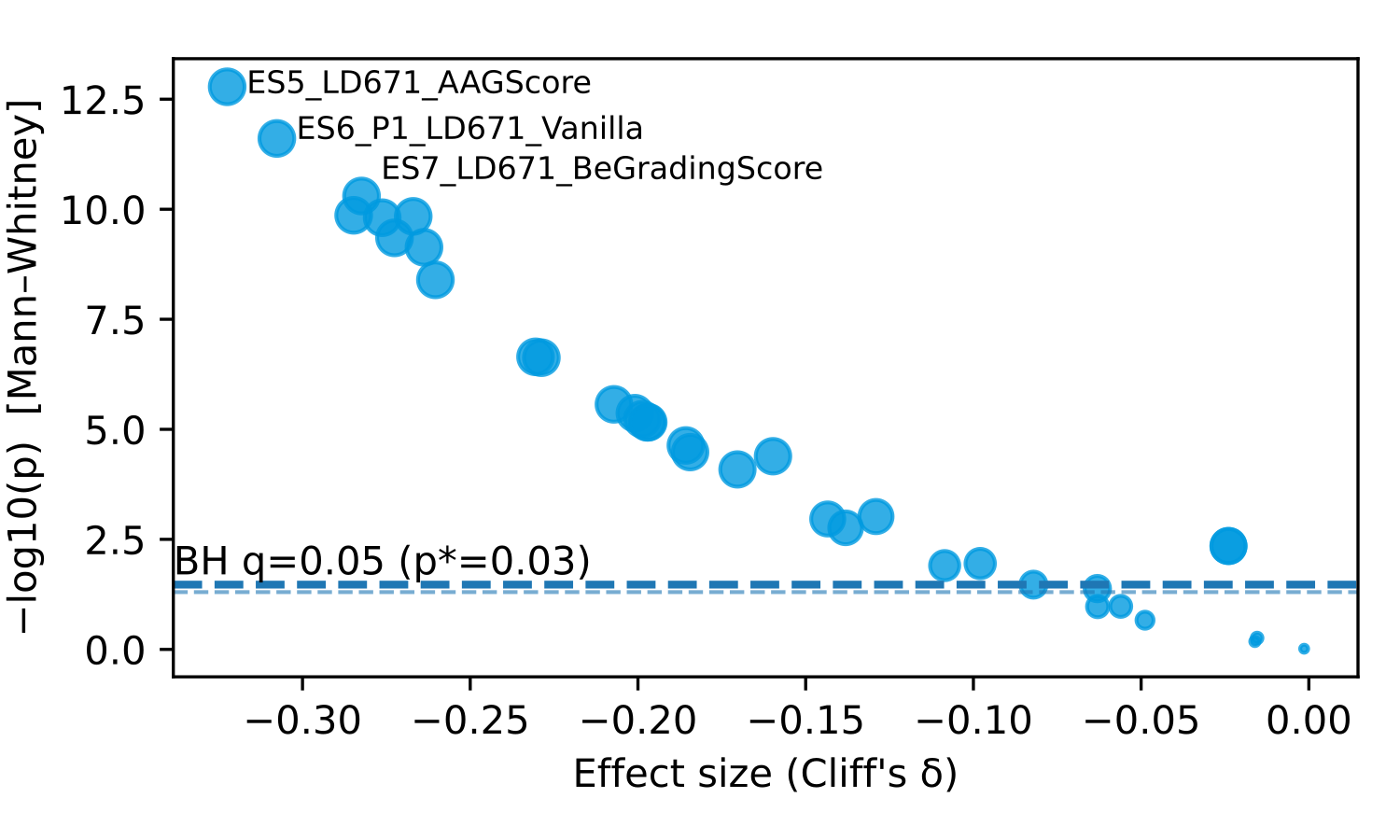}
    \caption{
    Volcano plot for DeepSeek models (same method as \autoref{fig:volcano_all}).
    }
    \label{fig:volcano_deepseek}
\end{figure}

Volcano plots were used to summarise the results (see \autoref{fig:volcano_all} for results on all models, and \autoref{fig:volcano_qwen}, \autoref{fig:volcano_qpt}, \autoref{fig:volcano_deepseek} showing per-model results), with each feature represented by a point at coordinates $(x,y) = (\delta_f, -\log_{10}(p_f))$.
The $x$-axis encodes the magnitude and direction of the effect, while the $y$-axis reflects the strength of evidence against the null hypothesis.
Points farther from $x=0$ indicate larger effect sizes, and points higher on the $y$-axis correspond to smaller $p$-values.
Horizontal dashed lines mark the nominal significance threshold ($p = 0.05$) and the BH-adjusted FDR threshold ($p^\star$).


\begin{figure}[!htbp]
    \centering
    \includegraphics[width=0.32\linewidth]{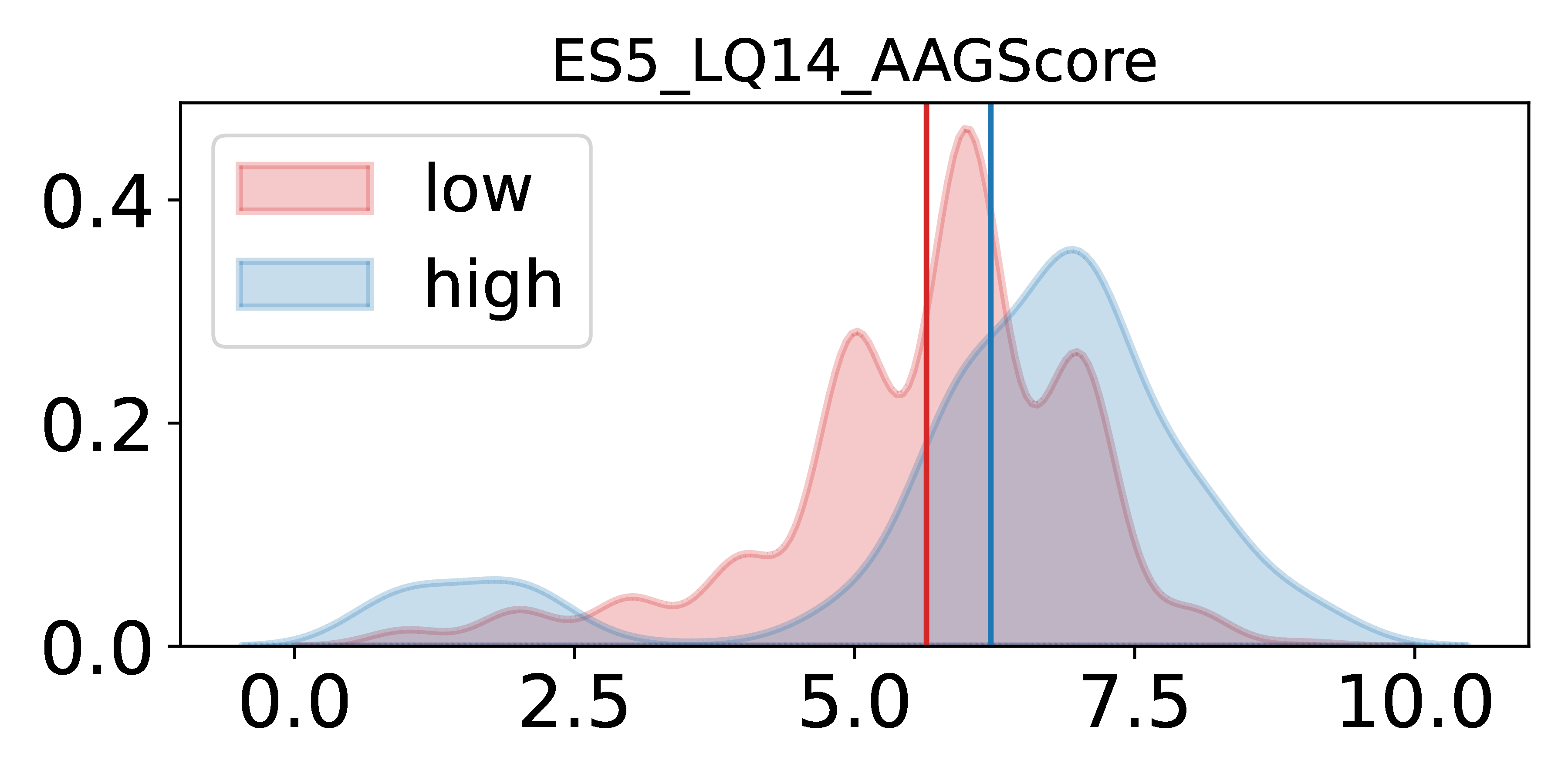}
    \includegraphics[width=0.32\linewidth]{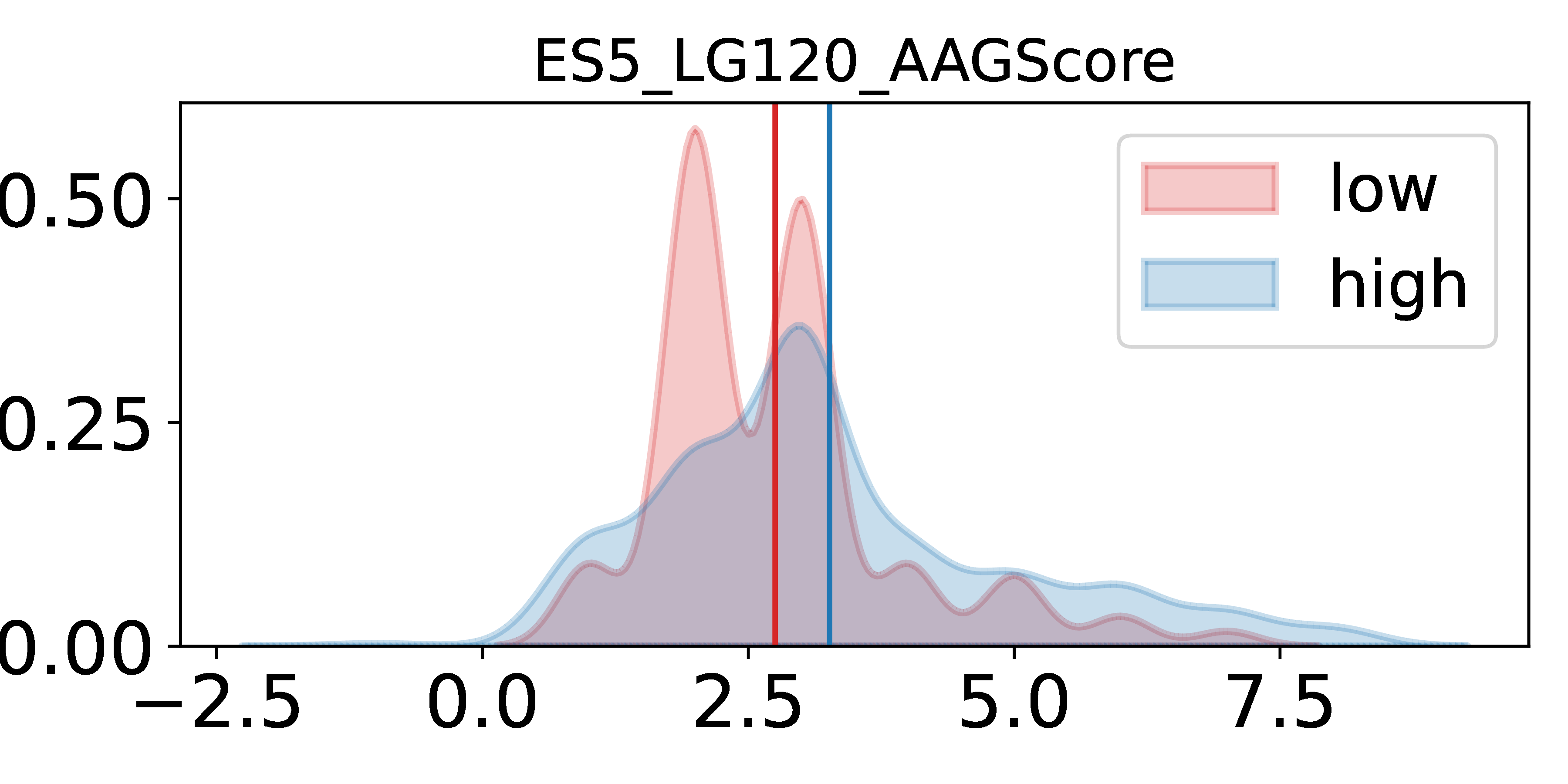}
    \caption{
    Kernel density estimates (KDEs) for two performance classes (low, high).
    }
    \label{fig:kde}
\end{figure}

\begin{figure}[!htbp]
    \centering
    \includegraphics[width=0.9\linewidth]{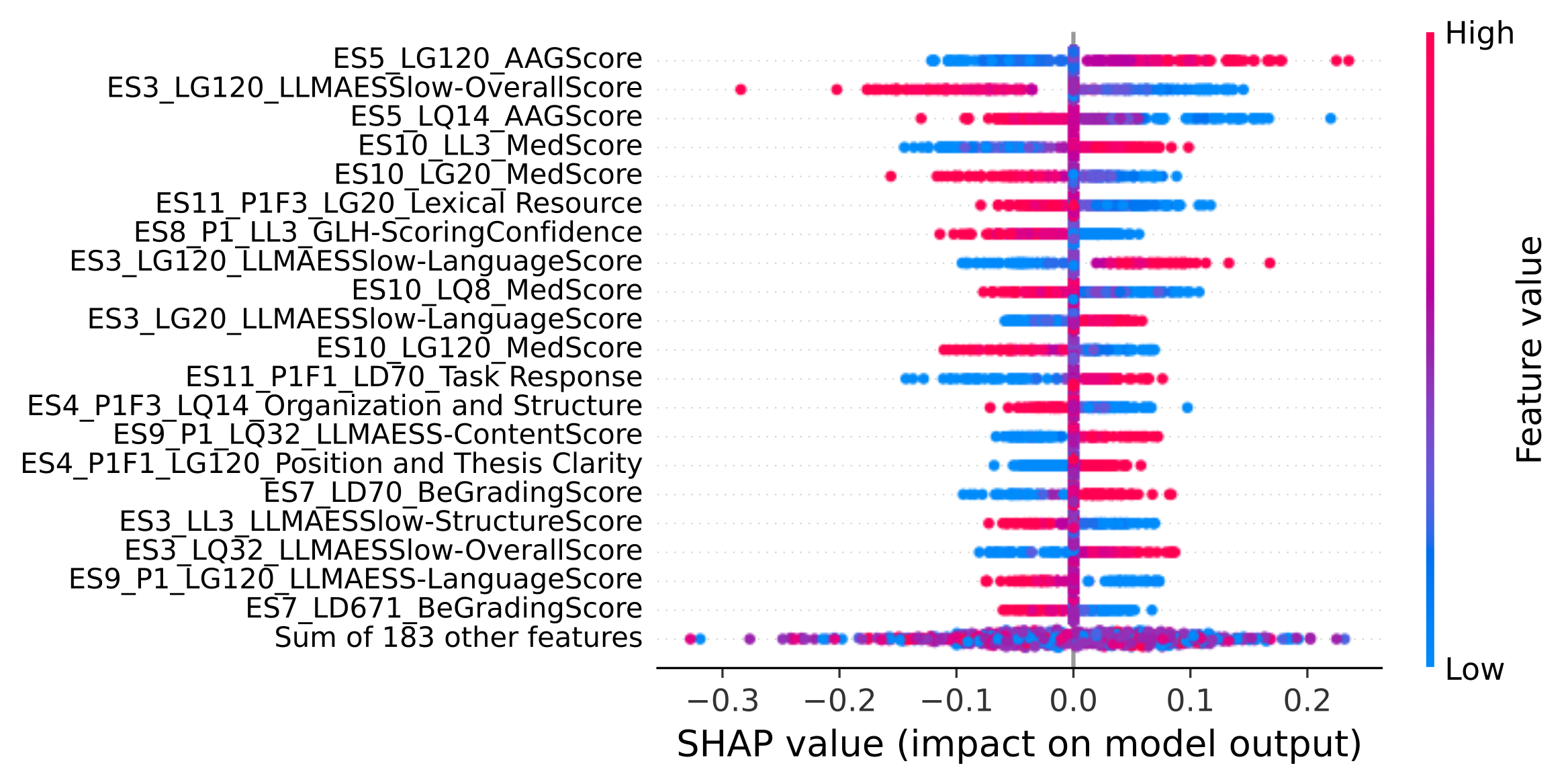}
    \caption{
    Global feature importance for \emph{low} performance class (quantile binning).
    }
    \label{fig:shap_beeswarm_global_quantile_0}
\end{figure}

\begin{figure}[!htbp]
    \centering
    \includegraphics[width=0.9\linewidth]{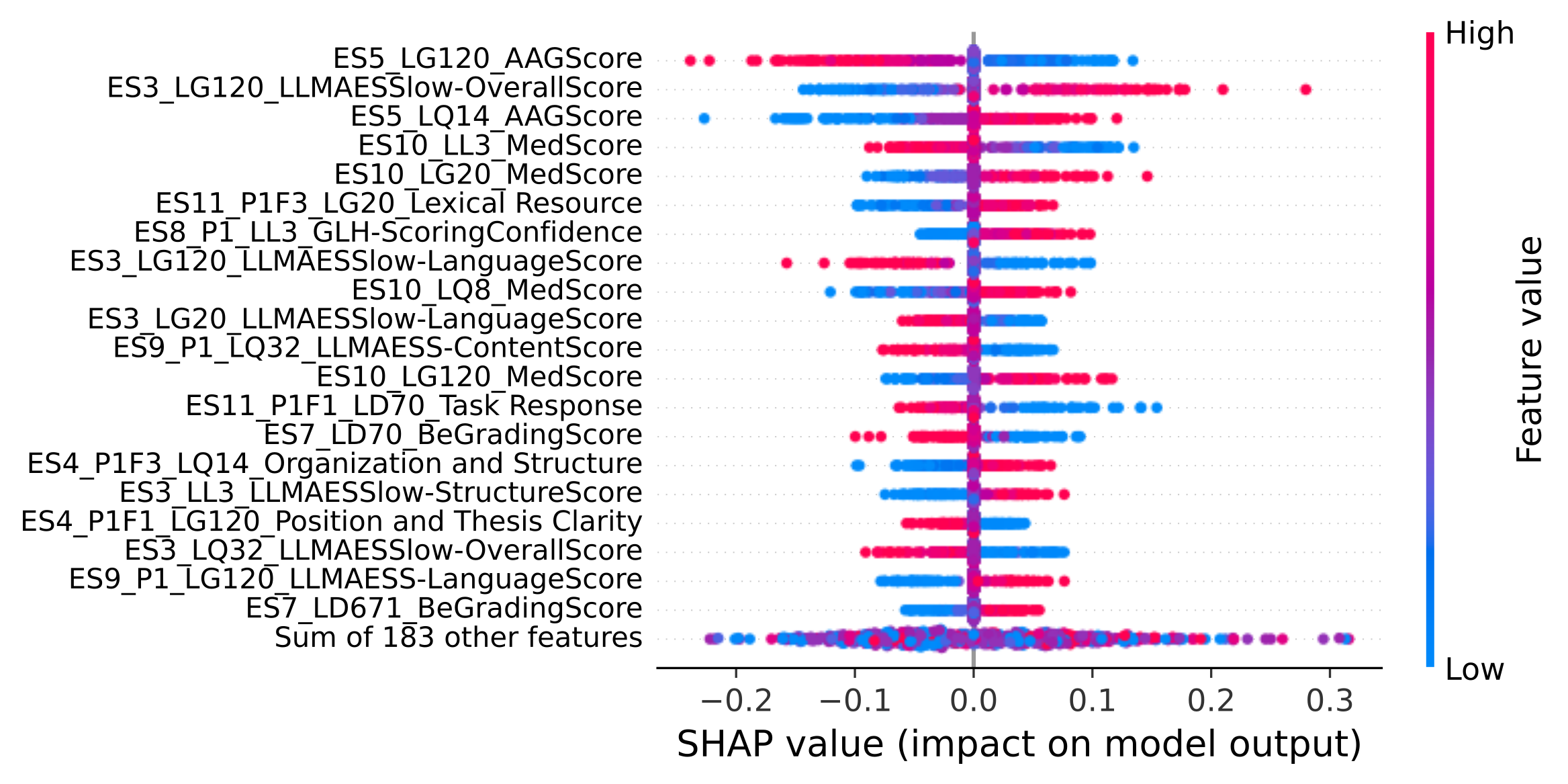}
    \caption{
    Global feature importance for \emph{high} performance class (quantile binning).
    }
    \label{fig:shap_beeswarm_global_quantile_1}
\end{figure}

\begin{figure}[!htbp]
    \centering
    \includegraphics[width=0.8
    \linewidth]{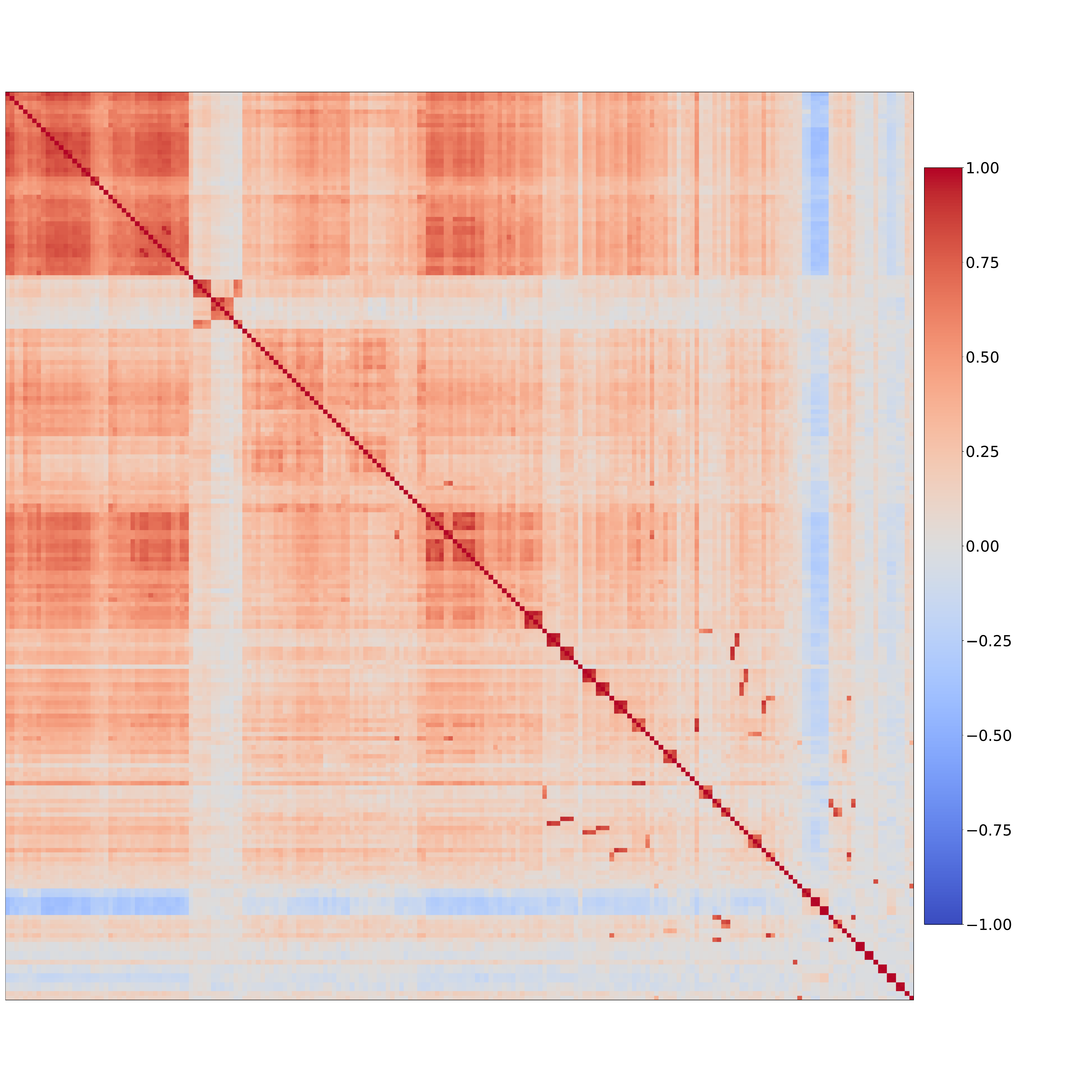}
    \caption{
    Pearson correlation matrix of feature vectors. The colour scale indicates pairwise linear correlations between features (1 = strong positive, -1 = strong negative, 0 = no linear relationship).
    }
    \label{fig:heatmap_feature_cor}
\end{figure}

In addition to the distribution analysis, we conducted a feature importance analysis using SHAP (SHapley Additive exPlanations) \cite{Lundberg:Lee:2017}.
The SHAP beeswarm plots visualise the global feature importance derived from the predictive model (linear SVM), showing how each feature contributes to the model output for the two focus classes (\emph{low} and \emph{high}), see \autoref{fig:shap_beeswarm_global_quantile_0} and \autoref{fig:shap_beeswarm_global_quantile_1}.
In each beeswarm plot, the features are ordered by their overall mean absolute SHAP value, with individual points representing samples.
The horizontal axis indicates the SHAP value, that is, the direction and magnitude of each feature's contribution to the prediction of the target class.
Colour coding is used to represent feature values. Blue and red represent low and high values, respectively.
Features positioned further to the right exert a stronger positive influence on the probability of the focus class, while features shifted left reduce it.
For the \emph{high} class, the SHAP distributions indicate that approximately half of the top features have higher values (red points) associated with positive SHAP contributions, increasing the model's decision function, that is shifting the prediction toward the \emph{high} class.
The corresponding SHAP plot for the \emph{low} class (\autoref{fig:shap_beeswarm_global_quantile_0}) is essentially its mirror image.
The remaining features show the inverse pattern, with red points concentrated at negative SHAP values, implying that higher feature values reduce the likelihood of the \emph{high} outcome.
Notably, the same feature annotated by different LLMs can display opposite associations.
For example, \textit{AAGScore} from \textit{LG120} (GPT-OSS:120b) and \textit{LQ14} (Qwen3:14b) show opposite SHAP patterns despite similar univariate distributions -- the \emph{high} class tends to receive higher scores from both (see \autoref{fig:kde}).
This reflects multicollinearity -- with overlapping predictors, the local (conditional) contribution of a feature can become negative even when its marginal association with the outcome is positive, because the attribution (and, in linear models, the coefficient sign) captures a partial effect after accounting for correlated covariates.
Consequently, a feature positively correlated with the target can receive a negative coefficient if it overlaps with other predictors that track the same outcome direction.
Linear models redistribute weights across correlated signals, so the raw correlation sign need not match the learned partial effect.
As shown in \autoref{fig:heatmap_feature_cor}, several feature clusters exhibit moderate to strong intercorrelations.
This is consistent with the fact that the automatic essay scoring systems are all intended to assess related aspects of overall writing quality (e.g., coherence, grammatical accuracy, and content organisation), and thus their scores are expected to be positively associated.
However, because each system operationalises writing quality differently and introduces its own measurement noise, we do not expect perfect agreement between them.
The largest cluster of features in this study, located in the upper-left quadrant of the heatmap, comprises 41 features with a mean within-cluster Pearson correlation of 0.66, indicating substantial internal consistency.
In contrast, the model confidence scores are approximately uncorrelated with the other features (mean Pearson correlation = –0.06), suggesting that scoring uncertainty captures information distinct from the linguistic and rubric-based measures.

\begin{figure}[!htbp]
    \centering
    \includegraphics[width=0.9\linewidth]{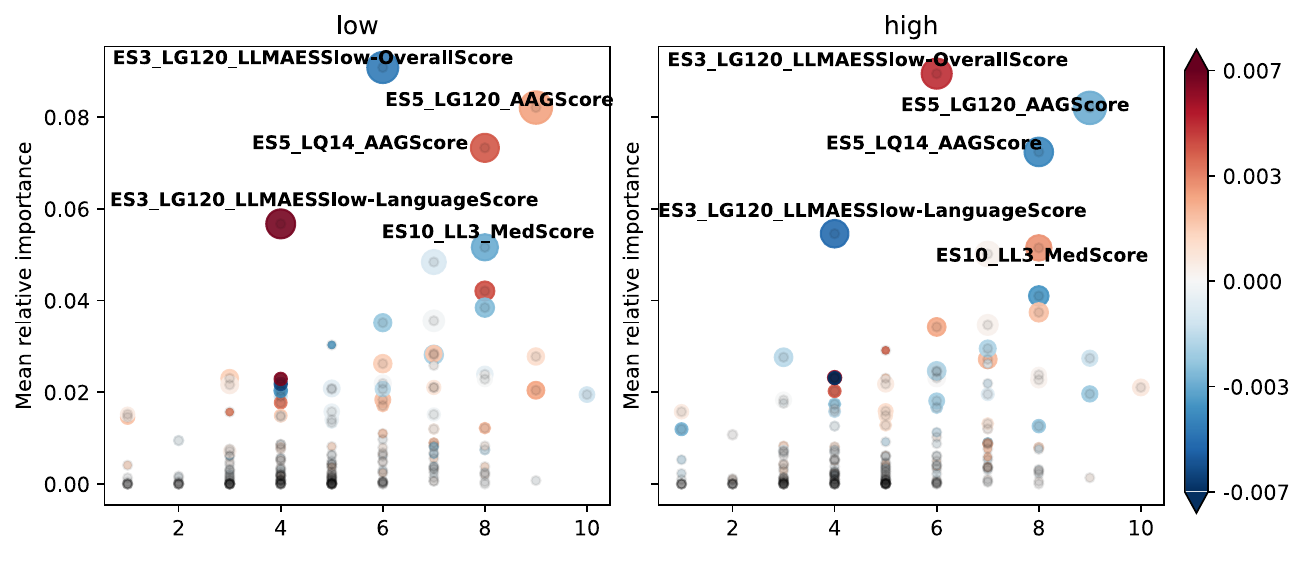}
    \caption{
    Feature stability and importance across models. The x-axis encodes the number of cross-validation splits (out of 10 total), where the feature is selected by the evolutionary search; the y-axis encodes the magnitude of SHAP importance, the colours indicate the direction (blue - negative, red positive).
    }
    \label{fig:shap_prevalence_global}
\end{figure}

Together, the volcano and SHAP analyses point to a consistent pattern: features related to computational correctness and explanatory clarity (\textit{AAGScore}) and to content accuracy and completeness (\textit{MedScore}), that is, \textit{features tied to content and reasoning rather than surface form}, most strongly separate the \emph{high} performance class from the others.
The volcano plots quantify these as statistically significant distributional differences, whereas the SHAP beeswarm plots show how the model integrates these mixed signals, amplifying coherent features while balancing contrasting contributions from different language-model sources.
\autoref{fig:shap_prevalence_global} further supports robustness of \textit{AAGScore} and \textit{MedScore}: across ten cross-validation splits, both features are repeatedly selected (8-9 times) and exhibit high mean relative SHAP importance, underscoring their stable contribution to the model's predictions.
Because the medical-education tasks from the CORE project target the \ERGS{} facet of reasoning and argumentation, our results indicate that the AES features most strongly capture reasoning- and content-related signals when predicting high-performing responses. 
This alignment with \ERGS{} suggests that our classifier reflects task-relevant constructs rather than superficial writing traits.
We can see an indication of the \ERGS{}-related nature of our AES-based classification here, which ultimately indicates some validity of our approach.


\section{Discussion}
\label{sec:discussion}
\begin{figure}[!htbp]
    \centering
    \includegraphics[width=.5\linewidth]{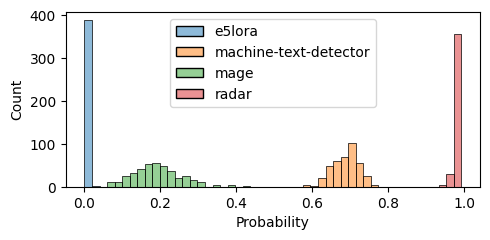}
    \caption{
    Histogram of 4 LLM-detection tools showing the probabilities of proband statements being LLM-generated based on the averaged per-sentence probability.
    }
    \label{fig:llm_detection_histogram}
\end{figure}

Our experiments show that a classifier trained on feature ensembles from multiple AES can reproduce student performance levels as defined by aggregated human ratings. 
This approach avoids reliance on a single scoring model and instead integrates complementary information across AES.
Because the classifier operates on feature vectors rather than predefined score scales, it can accommodate different performance-class configurations.
We illustrate this using four class variants, analysing the differences in performance classes on a relative and absolute scale using the quantile and uniform binning strategy.
Across all settings, our classifier achieves high Macro F1 scores with only about 15\% of the available features (see \autoref{fig:results:evolsearch:all_llms}), demonstrating that accurate predictions do not require the full feature set.

Although the AES outputs strongly correlate (see \autoref{fig:heatmap_feature_cor}) and the results of the different LLMs are mostly comparable (see Figures~\ref{fig:results:evolsearch:2_quantile_ds_gpt} and~\ref{fig:results:evolsearch:2_quantile_2_llama_qwen}), combining the features of different tools and LLMs increases the performance in all cases.
The distribution analysis shows that the most predictive features differ across LLMs, and that 7 of the 11 AES contribute at least one top-ranked feature. 
We also observe that LLM parameter size does not affect performance.
This highlights the influence of prompt design in AES and shows that our classifier can partially compensate for weaker individual LLMs by leveraging complementary information across models.
This is evident in the higher F1 scores of these groups and in the list of the most important features by SHAP values, in which all LLMs are among the top 10\%.

The SHAP analysis shows the frequent occurrence of \textit{MedScore}, which indicates a certain level of specialization in the field. 
This raises the question of whether our findings can be transferred to other domains.
However, as noted in \autoref{sec:methods:essay_scorer}, the only domain-specific term in the \textit{ES10} system prompt was \textit{medical teacher}. We replaced it with the more general hypernym \textit{teacher} to remove the explicit medical cue.
Across all binning variants, the \textit{MedScore} feature was selected in every of the 10 evolutionary-search splits and appeared in the top features of four to five LLMs on average.
Using only the \textit{MedScore} feature type (one feature per LLM, yielding eight input features), the classifier reaches 68\% Macro F1. This is lower than the 80\% achieved when the evolutionary search was allowed to select up to 10 features across all LLMs.
In contrast, removing the \textit{MedScore} feature, yielding 195 input features, does not reduce performance: the classifier still achieves about 91\% Macro F1.
These findings suggest that the output features of ES10 can be replaced with features from other non-domain-specific AES.
Although our approach is designed to be domain-independent and uses AES not specialised for medicine, our evaluation demonstrates generalization only across different medical assessment tasks.
Thus, future research should examine this generalizability in more detail.

Our findings also show that manually created gold data is still necessary to improve prediction accuracy and to assess AES reliability.
This issue is particularly relevant given the growing use of LLMs by students to support their writing. 
To examine whether AES can reliably grade responses that may contain LLM-generated content, we applied four LLM-detection tools~\cite{Hu:et:al:2023,Dugan:et:al:2024,Li:et:al:2024,Drayson:et:al:2025} to all student texts.
\autoref{fig:llm_detection_histogram} shows a strong disagreement among the detectors regarding whether the student responses were LLM-generated.
In light of these contradictory results, we must assume that the answers were written by humans.
However, further analysis of AES behaviour regarding LLM-generated texts is required and should be another focus of future research.

Although manually rated examples remain necessary, our results show that neither expensive LLM fine-tuning nor few-shot prompting is required, which keeps prompts short and reduces complexity. 
High output quality and efficient performance are especially important for real-time scoring during the writing process, where timely, personalised feedback becomes feasible. 
Because feature generation with LLMs is slow and costly, adopting smaller and faster models could further improve efficiency. 
In any event and consistent with the SHAP and distribution analyses, neither of the transformer-based AES examined here contributed strongly to the classification performance.

Our results do not falsify H1: 
The AES and LLMs examined here provide features that allow a classifier to approximate aggregated expert ratings of student texts. 
Under the conditions of the CORE data setting (\autoref{sec:introduction}), the combined AES feature ensemble supports classification performance that is not attainable with any single AES alone. 
This suggests that an ensemble-based approach can generalise to similar text-based assessment tasks and may reduce, though not replace, the need for human rating. 
Our main methodological insight is threefold:
\begin{enumerate}
    \item performance relies on an optimised combination of AES, not on any single system;
    \item a simple linear SVM suffices to exploit these features; and
    \item domain-specific medical tuning is not required: \textit{general-purpose LLMs are sufficient}.
\end{enumerate}
The targeted annotation performance is therefore within reach using the machine learning approach we have tested. 
Whether this also applies to related areas of medical education and other fields, such as economics and physics, is a question for future research.

\section{Conclusion}\label{sec:conclusion}
We evaluated LLM-based AES systems in the context of medical education, using student responses from critical online reasoning tasks. Our findings show that combining features from multiple AES systems enables training a classifier that closely approximates expert ratings. The approach is modular and accommodates different performance class configurations without requiring prompt modifications or fine-tuning of the underlying LLMs.

Learner classes, defined by aggregated human ratings, could be reconstructed with high accuracy using a simple linear support vector machine (SVM) trained on a small set of AES-based features. The classifier's performance was consistent across domains and tasks, demonstrating strong generalizability. Features related to reasoning and content were the most predictive, while domain-specificity played a minor role.

This suggests that out-of-the-box LLMs, even without domain-specific fine-tuning, contain sufficient information for automatic essay grading in medical education. However, individual AES systems alone are insufficient; only their combination yields accurate predictions approximating human evaluators.

These findings point to a reduced need for manual scoring: automated scoring could partially replace or effectively support human evaluation.
However, manually rated examples remain essential for training and evaluating the classifier, and zero-shot AES prompts likewise require empirical validation.
Future work will investigate the generalizability of our approach to other domains and incorporate a broader range of scoring models.

\section*{Conflict of Interest}
\label{sec:conflict_interest_statement}

The authors declare no potential conflict of interests.

\section*{Acknowledgments}
\label{sec:acknowledgements}

This work was supported by the research group \emph{Critical Online Reasoning in Higher Education} (CORE, FOR 5404, grant number 462702138) sub-projects B05 and C08, and the Infrastructure Priority Programme \emph{New Data Spaces for the Social Sciences} (SPP 2431, grant number 539634240) sub-project ENTAILab, both funded by the German Research Foundation (DFG).

\printbibliography

\end{document}